\documentclass{article} 
\usepackage{iclr2027_conference,times}
\usepackage{enumitem}

\usepackage{amsmath,amsfonts,bm}

\def\eqref#1{equation~\ref{#1}}

\def\1{\bm{1}}

\DeclareMathAlphabet{\mathsfit}{\encodingdefault}{\sfdefault}{m}{sl}
\SetMathAlphabet{\mathsfit}{bold}{\encodingdefault}{\sfdefault}{bx}{n}

\usepackage{hyperref}
\usepackage{url}
\usepackage{graphicx}
\usepackage{wrapfig}
\usepackage{booktabs}
\usepackage[table]{xcolor}
\usepackage{multirow}
\usepackage{cleveref}
\usepackage{subcaption}
\usepackage{algorithm}
\usepackage{algpseudocode}

\usepackage{tabularx}

\IfFileExists{tables/prior_scores.tex}
  {\input{tables/prior_scores.tex}}{}

\definecolor{baseGreen}{HTML}{087F6D}

\usepackage{booktabs}
\usepackage{xcolor}

\definecolor{prcBlue}{HTML}{2463A6}

\newcommand{\auxbest}[1]{\textbf{#1}}

\newcommand{\auxgain}[2]{%
  \shortstack[r]{#1\\\textcolor{prcBlue}{(#2)}}%
}

\newcommand{\noisetrajpanel}[2]{%
  \begin{minipage}[t]{0.49\linewidth}
    \centering
    \includegraphics[
      width=\linewidth,
      height=0.18\textheight,
      keepaspectratio
    ]{#1}
    \par\smallskip
    {\scriptsize #2\par}
  \end{minipage}%
}

\title{Deep Prior Learning for Embodied Perception} 

\author{Yimou Wu, Jiaxin Guo, Yun-hui Liu\textsuperscript{*} \& Zheng Li\textsuperscript{*} \\
The Chinese University of Hong Kong\\
\texttt{\{yimouwu,jiaxinguo\}@link.cuhk.edu.hk} \\
\texttt{yhliu@mae.cuhk.edu.hk}, \texttt{lizheng@cuhk.edu.hk}
}

\iclrfinalcopy 
\begin{document}

\maketitle
\begingroup
\renewcommand\thefootnote{*} 
\footnotetext{Corresponding authors.}
\endgroup

\begin{figure}[h]  
  \centering
  \vspace{-2em}
  \includegraphics[width=\textwidth]{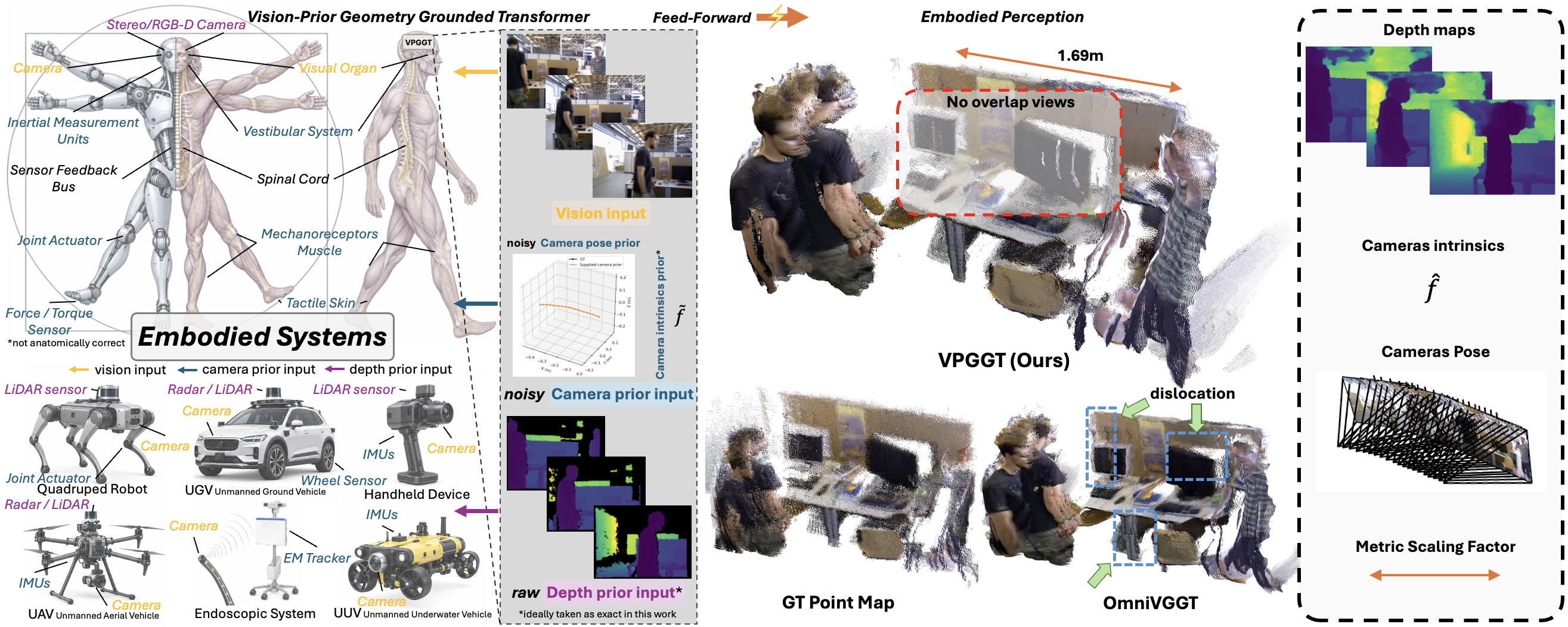}
  \caption{\textbf{VPGGT} is an embodied perception foundation model that combines images with optional camera parameters and depth observations to predict camera parameters, depth maps, and pointmaps with a shared metric scale. By leveraging geometric priors supplied by upstream sensing or estimation, the model circumvents the direct processing of raw sensor streams.}
  \label{fig:Overview}
\end{figure}

\begin{abstract}
Embodied systems need geometric perception that exploits available
observations beyond images alone. Recent feed-forward 3D models
incorporate geometric priors, including camera poses, intrinsics, and
depth. However, handling noisy poses, preserving accurate priors, and
recovering physical scale require more than simply accepting these
inputs. We introduce \emph{Vision-Prior Geometry Grounded Transformer}
(VPGGT), a VGGT-based framework that extends OmniVGGT for prior-aware
embodied perception. We formulate sensor-motivated pose corruptions from ground-truth trajectories for training and introduce a parameter-free \emph{prior residual connection} (PRC) to mitigate \emph{prior dilution}, where predictions are less accurate than their supplied pose priors. Our noise formulation targets camera poses; supplied intrinsics and depth receive no additional corruption. We further introduce \emph{Metric Global Attention}, which conditions a global
scale token on available pose and depth scales and predicts a shared
metric scaling factor for the geometric outputs. Experiments across four datasets show that \emph{PRC} improves translation-direction accuracy and joint pose AUC over a matched training baseline when camera priors are provided for all views, under both exact and corrupted poses. These results support explicit prior access during refinement as a useful addition to feature-level conditioning.
\end{abstract}


\section{Introduction}
\label{sec:intro}
We study feed-forward 3D perception from a set of images and optional
camera or depth priors (Fig.~\ref{fig:Overview}). DUSt3R~\citep{wang2024dust3r} and VGGT~\citep{Wang_2025_CVPR} have advanced image-based reconstruction by predicting multiple geometric attributes within a shared network. Their image-only inputs, however, do not exploit the camera estimates and depth observations often available to embodied systems. Pow3R~\citep{jang2025pow3r}, OmniVGGT~\citep{peng2026omnivggt}, and MapAnything~\citep{keetha2025mapanything} extend feed-forward reconstruction to incorporate such geometric information alongside images, providing a common model for different input configurations.

Several challenges remain when these models are used with observations
from embodied platforms. First, camera priors are estimates rather than ground truth, and their errors can accumulate over a trajectory or vary independently between frames. Although previous work has examined incorrect camera inputs, training and evaluation under structured pose errors require further study. Second, a model may produce camera estimates that are less accurate than the supplied priors, even when those priors are exact. We refer to this shortfall as \emph{prior dilution} (Sec.~\ref{subsec:prior_dilution}). Third, embodied systems require geometry in physical units, while normalizing geometric inputs removes their original scale. MapAnything~\citep{keetha2025mapanything} addresses metric reconstruction with a factored representation. Here, we study scale recovery while retaining the camera and dense-prediction heads of a pretrained VGGT-derived model.

In this paper, we introduce \emph{Vision-Prior Geometry Grounded
Transformer} (VPGGT), a feed-forward model that combines images with
optional camera and depth priors. First, we formulate sensor-motivated pose corruptions from existing ground-truth trajectories for training
(Sec.~\ref{subsec:noisy_prior_formulation}).\footnote{ We corrupt camera poses only. Supplied intrinsics and depth receive no additional corruption in this work.} Second, we introduce a parameter-free \emph{Prior Residual Connection} (PRC) that places scale-aligned priors in the camera head's refinement state to mitigate prior dilution (Sec.~\ref{subsubsec:prior_residual_connection}). Third, \emph{Metric Global Attention} conditions a global token on available pose and depth scales and predicts a shared metric factor (Sec.~\ref{subsubsec:metric_global_attention}). The model retains the image-only setting when geometric priors are absent and a common interface for further platform-specific training.

Comprehensive experiments and systematic architecture ablations examine the effects of prior quality and availability on camera and
depth prediction. Under matched training configurations, \emph{PRC} improves
translation-direction accuracy and joint pose AUC on all four datasets
when camera priors cover every view, for both exact and corrupted
poses. The improvements are largest on TUM RGB-D and C3VD, while results at partial coverage reveal remaining limitations. We study these effects separately from metric-scale recovery to distinguish better camera refinement from better estimation of physical scale.

\section{Related Work}
\label{sec:relatedwork}
\paragraph{Vision-Based 3D Reconstruction}
Traditional reconstruction systems combine correspondence estimation,
camera recovery, and sparse or dense mapping through task-specific
modules~\citep{schonberger2016structure,furukawa2010towards,campos2021orb}. Learned features, matchers, depth networks, and differentiable
optimization improve these components\citep{detone2018superpoint,sarlin2020superglue,sun2021loftr,
teed2021droid,yao2018mvsnet,wang2024vggsfm}. VGGSfM~\citep{wang2024vggsfm}, for example, integrates learned
correspondences and camera estimation with differentiable bundle
adjustment. In contrast, DUSt3R~\citep{wang2024dust3r} predicts dense pointmaps from image pairs, while VGGT~\citep{Wang_2025_CVPR} jointly predicts camera parameters and dense geometry from multiple views. These models reduce the need for separately designed reconstruction
stages, but their image-only interfaces do not use external camera
or depth observations.

\paragraph{Embodied Perception}
Embodied platforms may combine cameras with IMUs, wheel encoders,
LiDAR, and depth sensors\citep{zhao2025resilient,frey2026advances,mao2025multisensory,cullen2023internal}. Visual--inertial and multi-sensor SLAM fuse their measurements through state estimation and geometric optimization
\citep{qin2018vins,campos2021orb}. Their observation models, uncertainty estimates, and calibration describe how individual sensors constrain the estimated motion. Prior-conditioned feed-forward models operate at a different level, taking geometric estimates rather than raw sensor measurements. Pow3R~\citep{jang2025pow3r}, OmniVGGT~\citep{peng2026omnivggt}, and MapAnything~\citep{keetha2025mapanything} incorporate camera and depth information into dense reconstruction, with MapAnything also predicting metric scale. Pow3R examines responses to incorrect camera parameters, and MapAnything identifies noise and uncertainty in geometric inputs as a remaining limitation. Our work studies structured pose corruption, preservation of informative camera priors during refinement, and scale conditioning within a pretrained multi-view model.

\section{Preliminaries and Problem Definition}
\label{sec:preliminary_and_problem}

\subsection{Preliminaries}
\label{subsec:preliminary}

Our implementation builds on OmniVGGT~\citep{peng2026omnivggt},
retaining the backbone and prediction heads inherited from
VGGT~\citep{Wang_2025_CVPR}. Each image $I_i$ is encoded into patch tokens $\mathbf t_i^I$, one camera token $\mathbf t_i^g$, and four register tokens $\mathbf t_i^R$. The backbone contains $L=24$ pairs of frame-wise and global self-attention blocks. OmniVGGT's GeoAdapters add normalized depth features to image tokens and camera features to camera tokens, with stochastic prior availability during training. Depth and pointmap heads predict dense geometry, while the camera head refines a nine-dimensional camera estimate over four iterations. We retain these components and add a camera-state prior connection and a metric-scale prediction branch.

\subsection{Prior Dilution}
\label{subsec:prior_dilution}

\begin{wrapfigure}[17]{r}{0.49\textwidth}
  \centering
  \vspace{-1.5em}
  \includegraphics[width=0.49\textwidth]{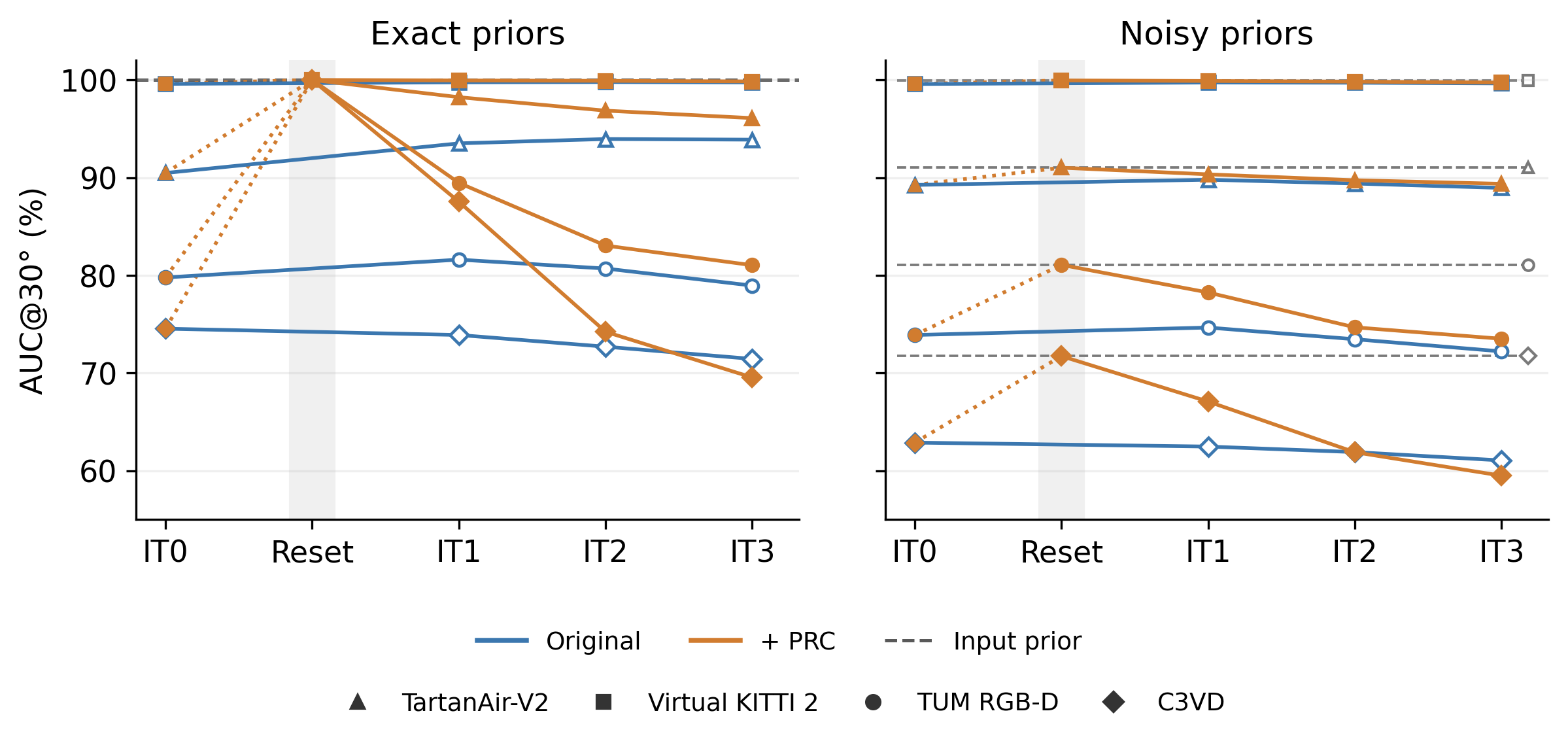}
  \caption{\textbf{Prior dilution during camera refinement.} A0 is released OmniVGGT; A1 uses the same checkpoint with \emph{PRC} enabled, without additional training. All views receive camera priors; no depth is supplied. Colors distinguish methods and markers distinguish datasets. Dashed lines show input-prior scores. Reset is a state intervention, not a learned iteration.}
  \label{fig:prior_dilution}
\end{wrapfigure}

We use \emph{prior} to denote a supplied geometric estimate. For a higher-is-better pose metric $M$, let $M_{\rm prior}$ and $M_{\rm out}$ be the scores of the input prior and model prediction on the same eligible image pairs. We define their gap as $d_M=M_{\rm prior}-M_{\rm out}$ and refer to $d_M>0$ as \emph{prior dilution}. When the prior outperforms the matched no-camera-prior reference $M_0$, we define
\begin{equation}
\kappa_M =
\frac{M_{\rm out}-M_0}{M_{\rm prior}-M_0},
\qquad
\mathcal D_M=1-\kappa_M.
\label{eq:closure}
\end{equation}
Here $\kappa_M$ measures closure of the baseline-to-prior score gap. For exact priors, the angular-score reference is 100. For noisy priors, $\kappa_M>1$ indicates an output better than the supplied estimate. At partial coverage, prior retention is evaluated only on provided--provided pairs, with all compared scores using the same pair support.

Figure~\ref{fig:prior_dilution} compares the released camera head with and without PRC, without additional training. The two variants have identical IT0 scores. Reset recovers the exact-prior score of 100 on all four datasets, but each subsequent refinement update reduces AUC@$30^\circ$. This effect is strongest on TUM RGB-D and C3VD, where A1 finishes at 81.03 and 69.54, respectively. On C3VD, the latter is also below the original model's 71.44. The same post-reset decline occurs with noisy priors. Thus, direct access to a prior state alone does not ensure that the existing refinement operator preserves its accuracy. We train the model with PRC in Sec.~\ref{subsubsec:prior_residual_connection}.

\section{Methodology} 
\label{sec:methodology}
VPGGT estimates scene geometry from images and optional camera and
depth priors. We describe its inputs and outputs in Sec.~\ref{subsec:overview}, followed by \emph{Prior Residual Connection} in Sec.~\ref{subsubsec:prior_residual_connection} and \emph{Metric Global Attention} in Sec.~\ref{subsubsec:metric_global_attention}. The training procedure is given in Sec.~\ref{subsec:training_procedure}.

\subsection{Overview of VPGGT}
\label{subsec:overview}

\begin{figure}[t]
  \centering
  \includegraphics[width=\linewidth]{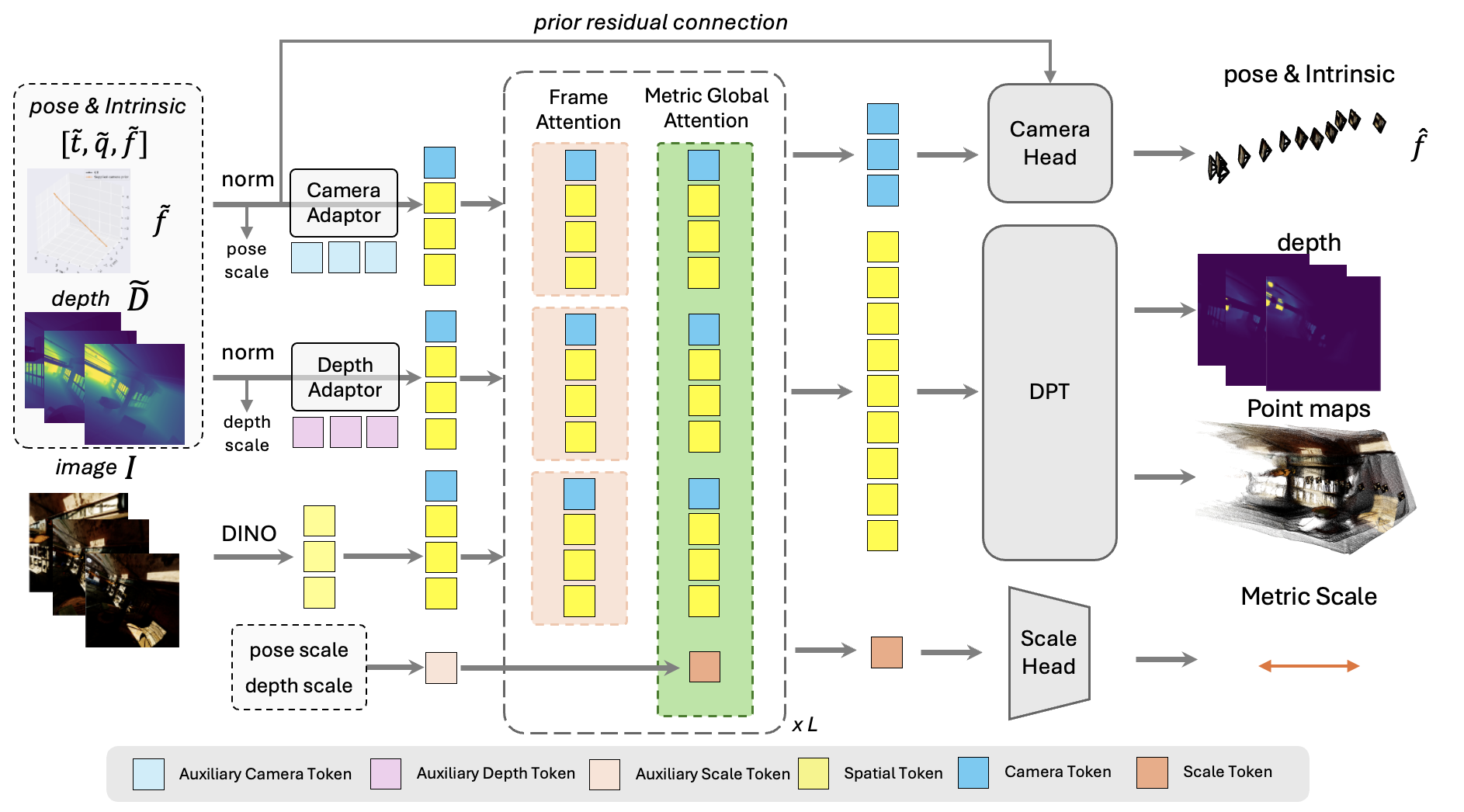}
  \caption{\textbf{Overview of VPGGT.} Images and optional camera or depth priors are encoded and processed by a shared alternating-attention backbone. \emph{PRC} places available camera priors in the camera head's refinement state after its initial prediction. A Scale Encoder conditions a global token on available metric-scale statistics, using learned placeholders when they are absent. The camera, depth, and pointmap heads predict geometry, and a scale head supplies a shared factor that converts translations, depths, and points to metric units.}
  \label{fig:architecture}
\end{figure}

Given images $\mathcal{I}=\{I_i\}_{i=1}^{N}$, the optional inputs comprise camera observations $\mathcal{C}=\{(K_j,G_j^{p})\}_{j\in\mathcal{Q}}$ and masked depth observations $\mathcal{D}=\{(D_k^{p},M_k)\}_{k\in\mathcal{O}}$, where $\mathcal{Q},\mathcal{O}\subseteq\{1,\ldots,N\}$. Camera poses use the world-to-camera convention, and a nonempty $\mathcal{Q}$ includes the reference view. The network predicts
\begin{equation}
f_\theta(\mathcal{I},\mathcal{C},\mathcal{D})
=
\left\{\hat{s},
(\hat{g}_i,\hat{D}_i,\hat{P}_i)_{i=1}^{N}\right\},
\label{eq:vpggt}
\end{equation}
together with depth and pointmap confidence maps. Here $\hat{g}_i=(\hat{\mathbf t}_i,\hat{\mathbf q}_i,\hat{\mathbf f}_i)
\in\mathbb{R}^{9}$ contains translation, a rotation quaternion, and two
fields of view. We retain the inherited camera model and first-view coordinate convention. The shared positive factor $\hat{s}$ produces
$\hat{D}_i^{\mathrm{met}}=\hat{s}\hat{D}_i$, $\hat{\mathbf t}_i^{\mathrm{met}}=\hat{s}\hat{\mathbf t}_i$, and $\hat{P}_i^{\mathrm{met}}=\hat{s}\hat{P}_i$; rotations and intrinsics are unchanged.

\subsection{Prior Residual Connection}
\label{subsubsec:prior_residual_connection}
\begin{wrapfigure}[27]{r}{0.35\textwidth}
  \centering
  \vspace{-1.3em}
  \includegraphics[width=0.35\textwidth]{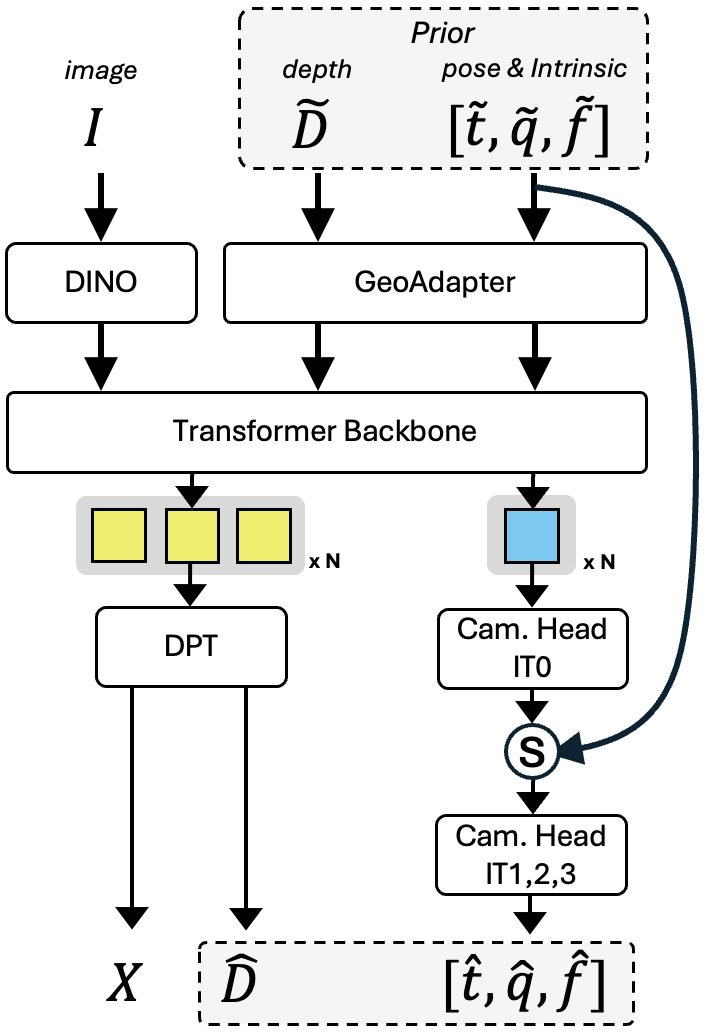}
  \caption{\textbf{Prior Residual Connection.} \emph{PRC} retains feature-level conditioning and supplies aligned camera priors as the state from which subsequent residual updates are predicted.}
  \label{fig:prc}
\end{wrapfigure}
The camera head already refines its predictions through residual
updates. \emph{PRC} changes the reference for these updates from an estimated camera state to the supplied prior wherever one is available
(Fig.~\ref{fig:prc}). This follows the residual-learning principle of learning corrections to a reference~\citep{he2016deep}, while retaining
the inherited camera-head architecture.

Let $\mathbf r_i^0$ denote the initial pre-activation camera estimate for view $i$. We align the normalized prior to its translation scale, obtaining $\mathbf a_i=(\alpha\bar{\mathbf t}_i^p,
\mathbf q_i^p,\mathbf f_i^p)$, where $\alpha$ is a translation norm ratio detailed in App.~\ref{app:prc_details}. After recording the initial prediction, \emph{PRC} sets
\begin{equation}
\mathbf u_i^0 =
\operatorname{sg}\!\left[
m_i\mathbf a_i+(1-m_i)\mathbf r_i^0
\right],
\label{eq:prc_reset}
\end{equation}
where $m_i$ indicates prior availability and $\operatorname{sg}$
denotes stop-gradient. The remaining iterations use the original refinement operator,
\begin{equation}
\mathbf u^k =
\operatorname{sg}(\mathbf u^{k-1})
+\Delta_\theta\!\left(
\mathbf H,\operatorname{sg}(\mathbf u^{k-1})
\right),
\quad k=1,\ldots,J-1,
\label{eq:prc_refinement}
\end{equation}
with camera-token features $\mathbf H$ and $J=4$ in our experiments.

For a provided view, the forward prediction thus consists of the aligned prior and the accumulated residual corrections. Zero corrections retain that prior state, while nonzero corrections allow the model to refine an imperfect input. \emph{PRC} adds no parameters and, for fixed network weights, leaves the original computation unchanged when camera priors are absent. It does not constrain the learned corrections to improve or preserve the input accuracy.

\begin{table}[t]
\centering
\caption{\textbf{Ablation of auxiliary information guidance.}
Depth and pose accuracy with varying prior coverage. \auxbest{Bold} identifies the better score between the matched-training OmniVGGT-trained and PRC-only models at identical input settings; released-model and input-prior references are not ranked. \textcolor{prcBlue}{Blue} parentheses report PRC-only gains over OmniVGGT-trained, in percentage points, with full camera priors and no depth input. Depth is median-aligned per frame; RTA and AUC use sign-free translation errors.}
\label{tab:auxiliary_guidance}

\begin{minipage}[t]{0.49\linewidth}
\centering
\scriptsize
\textbf{(a) TUM RGB-D}\\[3pt]
\setlength{\tabcolsep}{0.8pt}
\begin{tabular*}{\linewidth}
{@{\extracolsep{\fill}}rrcrrrrr@{}}
\toprule
\multicolumn{3}{c}{Priors}
& \multicolumn{2}{c}{Depth}
& \multicolumn{3}{c}{Camera} \\
\cmidrule(lr){1-3}\cmidrule(lr){4-5}\cmidrule(l){6-8}
D & C & Pose
& AbsRel$\downarrow$
& $\delta_1\uparrow$
& RRA@5$\uparrow$
& RTA@5$\uparrow$
& AUC@30$\uparrow$ \\
\midrule
\multicolumn{8}{l}{\textbf{\textit{Input prior reference}}} \\
100 & 100 & GT & 0.000 & 100.00 & 100.00 & 100.00 & 100.00 \\
100 & 100 & N  & 0.000 & 100.00 & 100.00 & 66.00 & 81.07 \\
\midrule
\multicolumn{8}{l}{\textbf{\textit{OmniVGGT, released}}} \\
0 & 0 & --
& .078 & 92.57 & 99.67
& 38.00 & 61.94 \\
100 & 100 & GT
& .018 & 99.20 & 97.22
& 67.78 & 76.66 \\
100 & 100 & N
& .018 & 99.20 & 97.22
& 57.11 & 71.30 \\
\midrule
\multicolumn{8}{l}{\textbf{\textit{OmniVGGT-trained}}} \\
0 & 0 & --
& .058 & \auxbest{94.64} & 96.33
& \auxbest{51.22} & 69.17 \\
0 & 100 & GT
& \auxbest{.057} & \auxbest{94.69} & \auxbest{97.22}
& 58.78 & 71.04 \\
0 & 100 & N
& \auxbest{.057} & \auxbest{94.70} & 97.22
& 58.89 & 70.89 \\
100 & 100 & GT
& .009 & 99.76 & 95.44
& 55.89 & 71.86 \\
100 & 100 & N
& .009 & 99.76 & 95.44
& 55.56 & 71.16 \\
\midrule
\multicolumn{8}{l}{\textbf{\textit{VPGGT, PRC only}}} \\
0 & 0 & --
& .058 & 94.41 & \auxbest{97.22}
& 50.33 & \auxbest{69.33} \\
\cmidrule(l){1-8}
30 & 0 & -- & .031 & 98.04 & 97.22 & 48.56 & 66.66 \\
50 & 0 & -- & .023 & 98.57 & 97.22 & 45.89 & 66.46 \\
70 & 0 & -- & .016 & 99.11 & 97.22 & 49.22 & 68.87 \\
100 & 0 & -- & .009 & 99.76 & 96.22 & 52.33 & 71.77 \\
\cmidrule(l){1-8}
0 & 30 & GT & .058 & 94.44 & 97.22 & 51.00 & 69.29 \\
0 & 50 & GT & .058 & 94.46 & 97.22 & 55.00 & 70.74 \\
0 & 70 & GT & .058 & 94.46 & 96.56 & 60.11 & 73.50 \\
0 & 100 & GT
& .058 & 94.49 & 96.89
& \auxgain{\auxbest{73.78}}{+15.00}
& \auxgain{\auxbest{80.46}}{+9.42} \\
\cmidrule(l){1-8}
0 & 30 & N & .058 & 94.44 & 97.22 & 49.78 & 68.64 \\
0 & 50 & N & .058 & 94.46 & 97.22 & 51.11 & 69.28 \\
0 & 70 & N & .058 & 94.46 & 96.56 & 55.00 & 70.89 \\
0 & 100 & N
& .058 & 94.51 & 97.22
& \auxgain{\auxbest{67.00}}{+8.11}
& \auxgain{\auxbest{76.37}}{+5.48} \\
\cmidrule(l){1-8}
100 & 100 & GT
& .009 & 99.76 & \auxbest{96.56}
& \auxbest{68.44} & \auxbest{79.08} \\
100 & 100 & N
& .009 & 99.76 & \auxbest{96.67}
& \auxbest{63.44} & \auxbest{75.31} \\
\bottomrule
\end{tabular*}
\end{minipage}
\hfill
\begin{minipage}[t]{0.49\linewidth}
\centering
\scriptsize
\textbf{(b) C3VD}\\[3pt]
\setlength{\tabcolsep}{0.8pt}
\begin{tabular*}{\linewidth}
{@{\extracolsep{\fill}}rrcrrrrr@{}}
\toprule
\multicolumn{3}{c}{Priors}
& \multicolumn{2}{c}{Depth}
& \multicolumn{3}{c}{Camera} \\
\cmidrule(lr){1-3}\cmidrule(lr){4-5}\cmidrule(l){6-8}
D & C & Pose
& AbsRel$\downarrow$
& $\delta_1\uparrow$
& RRA@5$\uparrow$
& RTA@5$\uparrow$
& AUC@30$\uparrow$ \\
\midrule
\multicolumn{8}{l}{\textbf{\textit{Input prior reference}}} \\
100 & 100 & GT & 0.000 & 100.00 & 100.00 & 100.00 & 100.00 \\
100 & 100 & N  & 0.000 & 100.00 & 100.00 & 57.73 & 71.73 \\
\midrule
\multicolumn{8}{l}{\textbf{\textit{OmniVGGT, released}}} \\
0 & 0 & --
& .511 & 29.89 & 87.51
& 14.67 & 40.88 \\
100 & 100 & GT
& .036 & 98.57 & 100.00
& 45.88 & 64.70 \\
100 & 100 & N
& .036 & 98.57 & 100.00
& 39.56 & 57.23 \\
\midrule
\multicolumn{8}{l}{\textbf{\textit{OmniVGGT-trained}}} \\
0 & 0 & --
& .055 & 97.87 & 100.00
& \auxbest{29.19} & \auxbest{61.55} \\
0 & 100 & GT
& .055 & 97.90 & 100.00
& 43.16 & 69.42 \\
0 & 100 & N
& .055 & 97.90 & 100.00
& 42.62 & 67.73 \\
100 & 100 & GT
& .004 & 99.98 & 100.00
& 55.51 & 76.36 \\
100 & 100 & N
& .004 & 99.98 & 100.00
& 54.07 & \auxbest{74.80} \\
\midrule
\multicolumn{8}{l}{\textbf{\textit{VPGGT, PRC only}}} \\
0 & 0 & --
& .055 & \auxbest{98.09} & 100.00
& 27.75 & 61.49 \\
\cmidrule(l){1-8}
30 & 0 & -- & .015 & 99.80 & 100.00 & 26.12 & 54.92 \\
50 & 0 & -- & .010 & 99.87 & 100.00 & 30.72 & 57.29 \\
70 & 0 & -- & .007 & 99.92 & 100.00 & 36.99 & 62.33 \\
100 & 0 & -- & .004 & 99.98 & 100.00 & 54.22 & 74.59 \\
\cmidrule(l){1-8}
0 & 30 & GT & .055 & 98.10 & 100.00 & 28.74 & 60.89 \\
0 & 50 & GT & .055 & 98.11 & 100.00 & 35.26 & 64.52 \\
0 & 70 & GT & .055 & 98.11 & 100.00 & 46.81 & 71.01 \\
0 & 100 & GT
& .055 & \auxbest{98.11} & 100.00
& \auxgain{\auxbest{76.74}}{+33.58}
& \auxgain{\auxbest{88.48}}{+19.06} \\
\cmidrule(l){1-8}
0 & 30 & N & .055 & 98.10 & 100.00 & 26.02 & 58.60 \\
0 & 50 & N & .055 & 98.11 & 100.00 & 28.84 & 58.77 \\
0 & 70 & N & .055 & 98.11 & 100.00 & 34.77 & 61.50 \\
0 & 100 & N
& .055 & \auxbest{98.11} & 100.00
& \auxgain{\auxbest{55.46}}{+12.84}
& \auxgain{\auxbest{72.81}}{+5.08} \\
\cmidrule(l){1-8}
100 & 100 & GT
& .004 & 99.98 & 100.00
& \auxbest{81.78} & \auxbest{91.04} \\
100 & 100 & N
& .004 & 99.98 & 100.00
& \auxbest{60.40} & 74.53 \\
\bottomrule
\end{tabular*}
\end{minipage}

\par\smallskip
{\scriptsize
D/C denote provided-view percentages.
GT/N specifies pose quality only; supplied depth is ground truth. Input-prior references copy ground-truth depth and score the supplied poses without a network.
Angular thresholds are in degrees. Ties at the displayed precision and configurations without a matched-training counterpart are unmarked. All scores except AbsRel are percentages.
\par}
\end{table}

\subsection{Metric Global Attention}
\label{subsubsec:metric_global_attention}
Metric Global Attention retains the physical-scale information removed when normalizing geometric priors (Fig.~\ref{fig:architecture}). Let $a_C$ and $a_D$ be the normalization factors computed from the supplied camera and depth observations. For each modality $j\in\{C,D\}$, a shared encoder maps its log scale and coverage $\rho_j$ to a feature. A learned modality-specific placeholder is used when the scale is missing or degenerate,\begin{equation}
\mathbf e_j =
\begin{cases}
E([\log a_j,\rho_j]), & v_j=1,\\
\mathbf p_j, & v_j=0,
\end{cases}
\qquad
\mathbf t_s^0 =
\mathbf t_{\rm learn}
+F_0([\mathbf e_C,\mathbf e_D,v_C,v_D]).
\label{eq:scale_conditioning}
\end{equation}
Here $v_j$ indicates a valid metric observation. The final projection of $F_0$ is zero-initialized, and all input scales are expressed in meters. Conditioning is applied once, before the attention blocks.

Following MapAnything~\citep{keetha2025mapanything}, the scale token
is appended to the multi-view sequence for global attention and bypasses frame attention. Its updated state is carried through all $L$ global blocks. The final token predicts a positive scale factor,
\begin{equation}
\hat s =
\exp\!\left(h(\operatorname{LN}(\mathbf t_s^L))\right).
\label{eq:scale_prediction}
\end{equation}
This factor scales the predicted depths, translations, and pointmaps
without changing their heads. When metric priors are absent, the same branch predicts scale from visual and normalized geometric features.

\subsection{Training Procedure}
\label{subsec:training_procedure}

\paragraph{Geometry and scale supervision.}
We retain the camera, depth, and pointmap objectives of VGGT and
OmniVGGT, with
$\mathcal L_{\rm geo}
=5\mathcal L_{\rm cam}
+\mathcal L_{\rm depth}
+\mathcal L_{\rm pmap}$.
Camera estimates receive $\ell_1$ supervision at all four refinement
iterations with iteration-dependent weights. Depth and pointmaps use confidence-weighted and unweighted regression, together with depth-gradient and pointmap-normal terms. These losses supervise the original geometric predictions before metric scaling.

The scale target is estimated from the current predictions and
metric ground truth. For depth, we use $\ell_D=\operatorname{median}_{p\in\Omega_D}
[\log D_p^*-\log\operatorname{sg}(\hat D_p)]$. For cameras, we use
$\ell_C=\operatorname{median}_{(i,j)\in\Omega_C}
[\log b_{ij}^*-\log\operatorname{sg}(\hat b_{ij})]$, where $b_{ij}$ is the distance between camera centers. The valid-modality mean defines $\ell^*$, giving $\mathcal L_{\rm scale}=|\log\hat s-\operatorname{sg}(\ell^*)|$. We optimize $\mathcal L_{\rm geo}+\mathcal L_{\rm scale}$,
omitting the scale term when no valid metric target is available. The target is detached from the geometric predictions, while the scale loss updates the scale branch and shared backbone.

\paragraph{Stochastic prior availability.}
We follow OmniVGGT's stochastic selection of camera and depth
observations. The Scale Encoder uses the same selected inputs and learned placeholders for unavailable scales. Metric ground truth supervises the scale prediction independently of input-prior availability, including image-only samples.

\section{Experiments}
\label{sec:experiment}

\paragraph{Datasets.}
We train VPGGT from the released OmniVGGT checkpoint using TartanAir~\citep{wang2020tartanair}, Virtual KITTI 2~\citep{cabon2020vkitti2}, TUM RGB-D~\citep{sturm2012benchmark}, and
C3VD~\citep{bobrow2023colonoscopy}. These datasets include simulated navigation and driving sequences, handheld RGB-D capture, and endoscopic imagery. Each epoch contains approximately 100k sampled training items. We evaluate on TartanAir-V2 and held-out sequences of the other three datasets, using ten-frame windows with stride three and
up to ten non-overlapping windows per sequence. Dataset splits, exact sampling budgets, and preprocessing details are provided in the appendix.

\paragraph{Pose-prior corruption.}
\label{subsec:noisy_prior_formulation}
\begin{wraptable}{r}{0.4\textwidth}
\centering
\vspace{-1.1em}
\setlength{\tabcolsep}{2.2pt}
\caption{Pose-corruption profiles. Drift is applied to relative-motion increments.}
\label{tab:noise}
\scriptsize
\begin{tabular}{@{}lll@{}}
\toprule
Training data & Translation & Rotation \\
\midrule
TartanAir & Time-dependent & Time-dependent \\
TUM RGB-D & Time-dependent & Time-dependent \\
Virtual KITTI 2 & Step-dependent & Time-dependent \\
C3VD & Independent & Independent \\
\bottomrule
\end{tabular}
\vspace{-0.9em}
\end{wraptable}
We generate corrupted camera trajectories in metric units before
view sampling and normalization. The drift profiles combine sequence-level biases with random motion increments, while the C3VD profile perturbs each pose independently (Table~\ref{tab:noise}). Each training item uses either the clean trajectory or one of four fixed corrupted versions, consistently across its views. Intrinsics and depth receive no additional corruption. The corruption models and parameters are given in App.~\ref{app:noisy_camera_pose_prior}.

\paragraph{Implementation Details.}
The backbone contains 24 frame/global attention pairs with token
dimension 1024. We retain the original geometric adapters and prediction heads. \emph{PRC} adds no parameters, while the scale branch adds approximately 0.447M parameters. For the prior-utilization study, we train the baseline and its \emph{PRC} variant from the same released OmniVGGT checkpoint for two epochs under matched configurations. This study used three NVIDIA H800 GPUs over approximately seven days. We then initialize VPGGT from the trained \emph{PRC} model and train the scale branch jointly with the network for two further epochs, using four H800 GPUs over approximately three days. Training uses AdamW, cosine learning-rate decay with warmup, bfloat16, gradient checkpointing, and gradient clipping at 1. Additional training and checkpoint settings are given in App.~\ref{app:training_details}.

\paragraph{Metrics.}
We report relative rotation accuracy (RRA), sign-free and directed translation-direction accuracy (RTA), and joint pose AUC. Joint errors are the per-pair maxima of rotation and translation errors; AUC uses the inherited one-degree histogram convention. Translation metrics exclude pairs with degenerate ground-truth baselines. Prior gaps $d_M$ and closures $\kappa_M$ are computed from the input-prior and prediction scores on identical support. Depth is evaluated both after scale alignment and in metric units; alignment settings are stated separately for each table.

\subsection{Ablation Study on Auxiliary Information Guidance}
\label{subsec:auxiliary_information_guidance}
We evaluate the effect of \emph{PRC} on held-out TUM RGB-D~\citep{sturm2012benchmark} and C3VD~\citep{bobrow2023colonoscopy} sequences. The baseline and PRC variant are initialized from the same released OmniVGGT~\citep{peng2026omnivggt} checkpoint and trained with matched data, optimization settings, and budgets. We denote them as OmniVGGT-trained and VPGGT (PRC only), respectively. Evaluation uses ten-frame windows at stride three, with images resized to width 518 and heights rounded to a multiple of 14. Camera and depth priors are supplied to the first 30\%, 50\%, 70\%, or 100\% of the views.

Table~\ref{tab:auxiliary_guidance} shows that \emph{PRC} improves camera-prior utilization, particularly at full coverage. With exact camera priors and no depth input, it raises RTA@$5^\circ$ from 58.78 to 73.78 on TUM RGB-D and from 43.16 to 76.74 on C3VD, gains of 15.00 and 33.58 percentage points. The corresponding AUC@$30^\circ$ gains are 9.42 and 19.06 points. With noisy poses, RTA improves by 8.11 and 12.84 points, while AUC improves by 5.48 and 5.08 points. The refined output also exceeds the noisy input reference on individual metrics: 67.00 versus 66.00 RTA@$5^\circ$ on TUM RGB-D, and 72.81 versus 71.73 AUC@$30^\circ$ on C3VD. These comparisons indicate improved refinement, but not uniform superiority over the supplied priors.

Depth observations mainly improve depth prediction. At 30\% depth coverage, the PRC variant reduces AbsRel from 0.058 to 0.031 on TUM RGB-D and from 0.055 to 0.015 on C3VD. Camera gains are less consistent at partial coverage. For example, at 30\% noisy camera coverage, C3VD RTA@$5^\circ$ decreases from 27.65 to 26.02 relative to OmniVGGT-trained. With full noisy camera and depth inputs, its C3VD AUC is also slightly lower, at 74.53 versus 74.80. \emph{PRC} therefore improves prior utilization without guaranteeing preservation under every input configuration. Complete results, stricter thresholds, directed metrics, and depth scores on unprovided views are reported in
App.~\ref{app:prc_full_results}.

\subsection{Camera Pose Estimation}
\label{subsec:camera_pose_estimation}

We compare full VPGGT with released OmniVGGT on TUM RGB-D, C3VD,
and fixed Sewerage and Ocean subsets of TartanAir-V2.
Released VGGT is additionally evaluated with RGB-only input on
the two TartanAir-V2 subsets.
All methods use the same ten-view windows with stride three:
20 windows from two TUM sequences, 45 from five C3VD sequences,
and 20 from four sequences in each TartanAir-V2 subset.
Sampling and scoring follow App.~\ref{app:evaluation_protocol}.
Full VPGGT includes PRC and MGA with additional fine-tuning;
these comparisons evaluate the resulting systems, not a
matched-training architectural change.

All settings include RGB.
We abbreviate exact and noisy camera inputs as
$C_e=(K,RT_{\rm exact})$ and $C_n=(K,RT_{\rm noisy})$.
Available priors cover every view.
Only poses are corrupted; intrinsics and supplied depth $D$
remain ground truth.
Exact-camera settings are therefore oracle-conditioned tests.

\begin{table}[t]
\centering
\scriptsize
\setlength{\tabcolsep}{2pt}
\renewcommand{\arraystretch}{1.0}
\caption{\textbf{Pose and depth on TUM RGB-D and C3VD.}
$R$, $T$, and $A$ denote RRA@$5^\circ$,
RTA@$5^\circ$, and AUC@$30^\circ$ in percent.
AR is median-aligned depth AbsRel;
$\delta_1$ is depth accuracy at threshold 1.25, in percent.
ATE uses Sim(3) alignment, in meters for TUM and millimeters
for C3VD.
Bold marks the unique best network within each input setting;
ties at displayed precision are unmarked.}
\label{tab:external_pose}
\label{tab:depth}
\label{tab:rotation_ate}
\begin{tabular*}{\linewidth}{@{\extracolsep{\fill}}l*{12}{r}@{}}
\toprule
& \multicolumn{6}{c}{TUM RGB-D}
& \multicolumn{6}{c}{C3VD}\\
\cmidrule(lr){2-7}\cmidrule(l){8-13}
Method
& $R\uparrow$ & $T\uparrow$ & $A\uparrow$
& AR$\downarrow$ & $\delta_1\uparrow$ & ATE$\downarrow$
& $R\uparrow$ & $T\uparrow$ & $A\uparrow$
& AR$\downarrow$ & $\delta_1\uparrow$ & ATE$\downarrow$\\
\midrule
\multicolumn{13}{l}{\textit{RGB only}}\\
OmniVGGT
& \textbf{99.67} & 38.00 & 61.94 & 0.0782 & 92.57 & 0.00554
& 87.51 & 14.67 & 40.88 & 0.5114 & 29.89 & 0.1962\\
VPGGT
& 96.22 & \textbf{50.44} & \textbf{69.38}
& \textbf{0.0607} & \textbf{94.21} & \textbf{0.00444}
& \textbf{100.00} & \textbf{27.41} & \textbf{62.58}
& \textbf{0.0563} & \textbf{97.86} & \textbf{0.0759}\\
\midrule
\multicolumn{13}{l}{\textit{w/ $C_e$}}\\
OmniVGGT
& \textbf{100.00} & 70.78 & 78.95 & 0.0779 & 92.18 & 0.00243
& 98.62 & 45.23 & 71.44 & 0.4848 & 30.96 & 0.0769\\
VPGGT
& 96.89 & \textbf{72.56} & \textbf{80.39}
& \textbf{0.0604} & \textbf{94.29} & \textbf{0.00223}
& \textbf{100.00} & \textbf{77.48} & \textbf{88.14}
& \textbf{0.0563} & \textbf{97.87} & \textbf{0.0210}\\
\midrule
\multicolumn{13}{l}{\textit{w/ $C_n$}}\\
OmniVGGT
& \textbf{100.00} & 58.89 & 72.21 & 0.0778 & 92.23 & 0.00263
& 98.67 & 36.35 & 61.05 & 0.4850 & 30.95 & 0.0873\\
VPGGT
& 97.11 & \textbf{67.44} & \textbf{76.50}
& \textbf{0.0604} & \textbf{94.29} & \textbf{0.00254}
& \textbf{100.00} & \textbf{56.05} & \textbf{72.95}
& \textbf{0.0563} & \textbf{97.87} & \textbf{0.0392}\\
\bottomrule
\end{tabular*}
\par\smallskip
\begin{minipage}{\linewidth}
\raggedright
Input-prior references:
exact $T/A=100/100$;
noisy $T/A=66.00/81.07$ on TUM and $57.73/71.73$ on C3VD.
Input-prior $R=100$ in both camera settings.
\end{minipage}
\end{table}

\begin{table}[t]
\centering
\scriptsize
\setlength{\tabcolsep}{2pt}
\renewcommand{\arraystretch}{1.0}
\caption{\textbf{Pose and depth on TartanAir-V2 subsets.}
Metrics and boldface follow Table~\ref{tab:external_pose};
ATE is in meters.
Sewerage uses P001/P002 and Ocean uses P000/P001
in both Easy and Hard, with five windows per sequence.
These results cover the selected subsets, not the full benchmark.}
\label{tab:tav2_tasks}
\begin{tabular*}{\linewidth}{@{\extracolsep{\fill}}l*{12}{r}@{}}
\toprule
& \multicolumn{6}{c}{Sewerage}
& \multicolumn{6}{c}{Ocean}\\
\cmidrule(lr){2-7}\cmidrule(l){8-13}
Method
& $R\uparrow$ & $T\uparrow$ & $A\uparrow$
& AR$\downarrow$ & $\delta_1\uparrow$ & ATE$\downarrow$
& $R\uparrow$ & $T\uparrow$ & $A\uparrow$
& AR$\downarrow$ & $\delta_1\uparrow$ & ATE$\downarrow$\\
\midrule
\multicolumn{13}{l}{\textit{RGB only}}\\
VGGT
& \textbf{94.67} & 67.44 & 84.51
& \textbf{0.0867} & \textbf{89.71} & 0.0892
& \textbf{91.22} & 42.78 & 75.51
& 0.0868 & 92.95 & 0.1786\\
OmniVGGT
& 88.56 & \textbf{69.11} & \textbf{84.63}
& 0.0940 & 87.92 & \textbf{0.0804}
& 90.67 & \textbf{45.00} & \textbf{78.14}
& \textbf{0.0777} & 94.89 & \textbf{0.1591}\\
VPGGT
& 83.33 & 53.56 & 78.84 & 0.1009 & 86.96 & 0.1148
& 90.33 & 42.78 & 75.68 & 0.0804 & \textbf{95.30} & 0.2079\\
\midrule
\multicolumn{13}{l}{\textit{w/ $C_e$}}\\
OmniVGGT
& \textbf{100.00} & \textbf{96.67} & \textbf{95.23}
& \textbf{0.0913} & \textbf{88.15} & \textbf{0.0408}
& 100.00 & \textbf{93.11} & \textbf{93.74}
& \textbf{0.0776} & 94.88 & \textbf{0.0557}\\
VPGGT
& 97.22 & 90.78 & 91.90 & 0.0977 & 87.25 & 0.0653
& 100.00 & 71.00 & 86.11 & 0.0791 & \textbf{95.57} & 0.1113\\
\midrule
\multicolumn{13}{l}{\textit{w/ $C_n$}}\\
OmniVGGT
& \textbf{100.00} & 85.00 & 90.24
& \textbf{0.0913} & \textbf{88.14} & \textbf{0.0472}
& \textbf{100.00} & \textbf{77.11} & \textbf{89.37}
& \textbf{0.0776} & 94.88 & \textbf{0.0636}\\
VPGGT
& 97.22 & \textbf{87.33} & \textbf{90.60}
& 0.0978 & 87.25 & 0.0690
& 99.89 & 70.56 & 85.81
& 0.0791 & \textbf{95.57} & 0.1127\\
\midrule
\multicolumn{13}{l}{\textit{w/ $D$}}\\
OmniVGGT
& 86.00 & \textbf{56.67} & \textbf{78.42}
& 0.0218 & 99.30 & \textbf{0.1147}
& \textbf{92.56} & \textbf{42.11} & \textbf{77.48}
& 0.0396 & 97.26 & \textbf{0.1587}\\
VPGGT
& \textbf{88.33} & 42.00 & 75.79
& \textbf{0.0105} & \textbf{99.64} & 0.1519
& 92.11 & 36.56 & 74.91
& \textbf{0.0258} & \textbf{98.40} & 0.1870\\
\midrule
\multicolumn{13}{l}{\textit{w/ $C_e,D$}}\\
OmniVGGT
& \textbf{97.78} & \textbf{94.22} & \textbf{95.01}
& 0.0210 & 99.30 & \textbf{0.0423}
& \textbf{100.00} & \textbf{93.78} & \textbf{94.36}
& 0.0394 & 97.29 & \textbf{0.0519}\\
VPGGT
& 96.89 & 84.56 & 88.87
& \textbf{0.0098} & \textbf{99.64} & 0.0754
& 99.44 & 81.89 & 88.31
& \textbf{0.0258} & \textbf{98.41} & 0.0954\\
\midrule
\multicolumn{13}{l}{\textit{w/ $C_n,D$}}\\
OmniVGGT
& \textbf{97.56} & \textbf{84.22} & \textbf{89.91}
& 0.0210 & 99.30 & \textbf{0.0526}
& \textbf{100.00} & 79.67 & \textbf{89.98}
& 0.0394 & 97.29 & \textbf{0.0598}\\
VPGGT
& 97.11 & 83.11 & 88.24
& \textbf{0.0098} & \textbf{99.64} & 0.0760
& 99.67 & \textbf{81.33} & 88.13
& \textbf{0.0258} & \textbf{98.41} & 0.0961\\
\bottomrule
\end{tabular*}
\par\smallskip
\begin{minipage}{\linewidth}
\raggedright
Noisy input-prior $T/A$:
Sewerage $86.78/90.90$; Ocean $82.78/91.66$,
with or without depth.
Exact-prior $T/A=100/100$; input-prior $R=100$.
\end{minipage}
\end{table}

VPGGT improves translation accuracy and joint pose AUC on
TUM/C3VD under all three settings, although TUM rotation
accuracy decreases (Table~\ref{tab:external_pose}).
The ranking largely reverses on TartanAir-V2:
OmniVGGT has higher AUC in 11 of 12 settings and lower
Sim(3)-aligned ATE in all 12 (Table~\ref{tab:tav2_tasks}).
VPGGT's sole AUC lead is on Sewerage with $C_n$
(90.60 versus 90.24).
Neither comparison establishes reliable prior improvement:
on Ocean with $C_n$, VPGGT obtains 85.81 AUC,
below the input prior's 91.66.

\subsection{Depth Estimation}
\label{subsec:depth_estimation}

We evaluate camera Z-depth from the same model executions,
using per-frame median alignment and shared ground-truth-valid
support, then average frame-level scores.
VPGGT reduces AbsRel on TUM/C3VD in all three settings.
Without depth inputs on TartanAir-V2, OmniVGGT instead has
lower AbsRel, while VPGGT retains higher $\delta_1$ on Ocean.
With depth inputs, VPGGT improves both metrics in all six
TartanAir-V2 comparisons.
These full-coverage settings assess preservation of observed
depth, not prediction of unavailable depth:
copying the supplied ground-truth depth would give zero
AbsRel and 100\% $\delta_1$ on the same valid pixels.

\subsection{Metric Reconstruction and Scale}
\label{subsec:scale_estimation}

VPGGT's predicted window scale is applied once to depth and
camera translations.
Metric AbsRel uses no median alignment, and ATE-SE(3) fits
rotation and translation only.
Following the point-norm scale statistic of
MapAnything~\citep{keetha2025mapanything}, we compute
\begin{equation}
E_{\rm scale}^{X}
=
100\left|
\frac{a(\hat X^{\rm met})}{a(X^*)}-1
\right|,
\qquad
a(X)=\frac{1}{|\Omega|}
\sum_{(i,p)\in\Omega}\|X_{i\rightarrow0}(p)\|_2.
\label{eq:metric_scale_eval}
\end{equation}
Each reconstruction is expressed in its own first-camera frame,
with common ground-truth-valid support $\Omega$.
Scale and trajectory errors are averaged over windows.
Because point radii also depend on shape and camera estimates,
this statistic does not isolate scale-head accuracy.

\begin{table}[t]
\centering
\scriptsize
\setlength{\tabcolsep}{2pt}
\renewcommand{\arraystretch}{1.05}
\caption{\textbf{Metric reconstruction of full VPGGT.}
Entries are metric AbsRel/$E_{\rm scale}^{X}$ (\%)/ATE-SE(3).
No post-hoc scale fitting to evaluation ground truth is used.
ATE is in millimeters for C3VD and meters for the other datasets.
All three metrics are lower-is-better; different input settings
are not ranked against one another.}
\label{tab:metric_scale}
\begin{minipage}[t]{0.49\linewidth}
\centering
\begin{tabular*}{\linewidth}{@{\extracolsep{\fill}}lcc@{}}
\toprule
Input & TUM RGB-D & C3VD\\
\midrule
RGB
& 0.1003/5.07/0.02226
& 0.0974/7.67/0.204\\
$C_e$
& 0.0977/4.88/0.02058
& 0.0900/6.80/0.173\\
$C_n$
& 0.0976/4.78/0.01811
& 0.0899/6.72/0.176\\
\bottomrule
\end{tabular*}
\end{minipage}\hfill
\begin{minipage}[t]{0.49\linewidth}
\centering
\begin{tabular*}{\linewidth}{@{\extracolsep{\fill}}lcc@{}}
\toprule
Input & Sewerage & Ocean\\
\midrule
RGB
& 0.1689/19.92/0.3626
& 0.1605/9.08/0.3522\\
$C_e$
& 0.1705/18.08/0.3482
& 0.1265/5.56/0.2349\\
$C_n$
& 0.1706/18.13/0.3498
& 0.1264/5.57/0.2346\\
$D$
& 0.1299/14.20/0.3329
& 0.0891/5.39/0.3589\\
$C_e,D$
& 0.1258/12.05/0.2726
& 0.0623/3.85/0.2312\\
$C_n,D$
& 0.1258/12.07/0.2734
& 0.0623/3.86/0.2305\\
\bottomrule
\end{tabular*}
\end{minipage}
\end{table}

RGB-only scale error ranges from 5.07\% on TUM to
19.92\% on Sewerage (Table~\ref{tab:metric_scale}).
Depth conditioning does not fully preserve physical scale:
on Sewerage with $C_e,D$, aligned AbsRel is 0.0098,
but metric AbsRel remains 0.1258.
Accurate aligned depth therefore does not imply accurate
metric reconstruction.

\paragraph{Limitations.}
The evaluation is restricted to small, fixed sequence subsets, so the reported gains on TUM/C3VD do not establish uniform cross-domain improvements. Pose corruption is synthetic, while intrinsics and supplied depth remain clean. The metric results only cover the full model, lacking a matched-training MGA ablation and an external metric baseline. A shared window scale cannot remove view-dependent distortions or enforce consistency across independently reconstructed windows. Single-frame inference, scene-level surface accuracy and completeness, clinical use, and closed-loop control are not evaluated.

\section{Conclusion}
\label{conclusion}
We present VPGGT, a feed-forward model for geometric perception from images and optional camera and depth priors. The model extends an OmniVGGT backbone with sensor-motivated pose-prior training, a parameter-free \emph{Prior Residual Connection}, and a prior-conditioned global scale token. \emph{PRC} gives the camera head direct access to supplied poses during refinement, while the scale branch converts the geometric outputs to metric units. Matched-training experiments show improved translation-direction accuracy and joint pose AUC under full camera-prior coverage, including corrupted poses. The results also show that prior availability alone does not ensure prior preservation, particularly at partial coverage. VPGGT provides a common model for incorporating geometric estimates into feed-forward perception and for further adaptation to the observations available on an embodied platform.

\subsection*{AI Use Statement}
We used AI tools to assist with manuscript drafting and editing,
discussion of evaluation protocols, and the drafting and debugging of
inference, visualization, and evaluation code.
Reported numerical results are obtained from model executions and
evaluation scripts, not generated by the language model.
The authors are responsible for the experimental decisions,
verification of the results and references, and the final manuscript.
\subsection*{Ethics Statement}
This work studies geometric perception using existing datasets and
synthetically corrupted camera estimates.
Its experiments do not establish the safety of deploying the model
for autonomous control or medical procedures.
Errors in camera pose, depth, or metric scale could cause harm if
used without independent validation and appropriate safeguards.
The endoscopic results should therefore be interpreted as benchmark
evaluations, not clinical validation.
Use of reconstruction systems must also account for privacy,
dataset permissions, and the possibility of recovering sensitive
information from captured environments.
\subsection*{Reproducibility Statement}
Section~\ref{sec:methodology} describes the model, and
App.~\ref{app:prc_details} specifies the PRC implementation.
Training and checkpoint settings are given in
App.~\ref{app:training_details}.
The pose-corruption process is described in
App.~\ref{app:noisy_camera_pose_prior}, and
App.~\ref{app:evaluation_protocol} defines window sampling,
input availability, validity masks, alignment, and aggregation.
Complete matched-training results are reported in
App.~\ref{app:prc_full_results}.
Experiment records retain checkpoint identifiers, source-window
identities, input conditions, and evaluation settings.

\bibliography{references}
\bibliographystyle{iclr2027_conference}

\appendix
\section{Appendix}



\subsection{Pose-Prior Corruption}
\label{app:noisy_camera_pose_prior}

We corrupt complete metric camera-to-world trajectories before
selecting training views or normalizing geometric inputs.
The profiles model errors in supplied camera estimates rather
than raw sensor measurements.
For each cached training realization, a sequence-level multiplier
$g\sim\mathrm{LogU}(0.6,1.8)$ scales the noise parameters.
Training selects either the clean trajectory or one of four cached
corrupted versions, using the same source for all views in a sample.

\paragraph{Accumulated drift.}
Let $T_k=[R_k,\mathbf c_k]$ denote a camera-to-world pose and
$\Delta T_k=T_{k-1}^{-1}T_k$ its relative motion.
Both drift profiles start from $T_0^p=T_0$ and accumulate
perturbed increments,
\begin{equation}
T_k^p =
T_{k-1}^p
\begin{bmatrix}
\delta R_k\Delta R_k &
\Delta\mathbf t_k+\delta\mathbf t_k\\
\mathbf 0^\top & 1
\end{bmatrix}.
\label{eq:noise_composition}
\end{equation}
Only the first pose of the complete trajectory is fixed.
The first pose of a later evaluation window need not be uncorrupted.

A sequence-constant heading-rate bias contributes
$b_\psi\Delta t_k$.
Angular white increments are scaled by
$q_k=\sqrt{\Delta t_k/t_0}$, where $t_0=0.1$\,s,
\begin{equation}
\delta\psi_k=b_\psi\Delta t_k+
\sigma_\psi q_k\epsilon_{\psi,k},
\qquad
\begin{bmatrix}\delta\theta_k\\\delta\phi_k\end{bmatrix}
=
\sigma_{rp}q_k
\begin{bmatrix}\epsilon_{\theta,k}\\\epsilon_{\phi,k}\end{bmatrix}.
\label{eq:noise_rotation}
\end{equation}
The random variables are standard normal.
All rotational components accumulate through trajectory composition;
no gravity-aiding correction bounds the other-axis errors.

The time-dependent translation profile uses a sequence-constant
velocity bias and white increments,
\begin{equation}
\delta\mathbf t_k =
\mathbf b_v\Delta t_k+
\sigma_t q_k\boldsymbol\epsilon_k.
\label{eq:noise_time_translation}
\end{equation}
The step-dependent profile instead uses
\begin{equation}
\delta\mathbf t_k =
b_s\Delta\mathbf t_k+
\left(
\sigma_f d_k+
\sigma_0\mathbf 1[d_k>5\sigma_0]
\right)\boldsymbol\epsilon_k,
\qquad
d_k=\|\Delta\mathbf t_k\|_2.
\label{eq:noise_distance_translation}
\end{equation}
Here $b_s$ is sampled once per trajectory.
The translation perturbation vanishes at an exactly stationary step,
but angular drift can continue.
The final trajectory error is not generally linear in time or
distance because rotational errors also affect subsequent translations.

\paragraph{Rotation axes.}
When gravity metadata is unavailable, the implementation estimates
a heading axis from the mean camera-up direction.
This direction is used when the norm of the mean exceeds 0.5;
otherwise, perturbations use a camera-axis Euler construction.
The procedure is a coordinate heuristic rather than a model of
gravity observability.

\paragraph{Independent perturbations.}
The tracker-style profile independently perturbs each absolute pose,
\begin{equation}
R_k^p=R_{\rm ZYX}(\boldsymbol\epsilon_{R,k})R_k,
\qquad
\mathbf c_k^p=\mathbf c_k+\boldsymbol\epsilon_{p,k}.
\label{eq:noise_independent}
\end{equation}
The angle and position components are Gaussian with the standard
deviations in Table~\ref{tab:noise_constants}.
The first pose is also perturbed.
The perturbations are independent conditional on the shared
sequence-level severity.
This is an idealized operating regime, not a calibration of a
particular tracking device.

\begin{table}[t]
\centering
\caption{Base pose-corruption parameters before severity scaling.
Time-scaled white increments use a reference interval of 0.1\,s.}
\label{tab:noise_constants}
\begingroup
\small
\setlength{\tabcolsep}{4pt}
\begin{tabular}{@{}llrrrr@{}}
\toprule
Parameter & Unit & TartanAir & TUM & VKITTI2 & C3VD \\
\midrule
Heading-rate bias std. & rad/s
& $4.37{\times}10^{-3}$ & $\pi/10800$ & $\pi/36000$ & -- \\
Heading white std. & deg
& 0.1100 & 0.08 & 0.03 & -- \\
Other-axis white std. & deg
& 0.02 & 0.03 & 0.02 & -- \\
Velocity-bias std. & m/s
& 0.076 & 0.0045 & -- & -- \\
Translation white std. & m
& 0.001 & 0.0009 & -- & -- \\
Translation scale-bias std. & --
& -- & -- & 0.01 & -- \\
Step-noise coefficient & --
& -- & -- & 0.01 & -- \\
Conditional translation floor & mm
& -- & -- & 0.5 & -- \\
Independent position std. & mm/axis
& -- & -- & -- & 0.03 \\
Independent Euler-angle std. & deg/axis
& -- & -- & -- & 0.05 \\
\bottomrule
\end{tabular}
\endgroup
\end{table}

\begin{algorithm}[t]
\caption{Accumulated pose-prior corruption}
\label{alg:drift}
\begin{algorithmic}[1]
\Require Camera-to-world poses $\{T_k\}$, time steps $\Delta t_k$,
profile parameters, severity $g$
\State Scale the profile parameters by $g$
\State Sample $b_\psi\sim\mathcal N(0,\sigma_{b\psi}^2)$
\If{time-dependent translation}
  \State Sample $\mathbf b_v\sim
  \mathcal N(\mathbf 0,\sigma_{bv}^2I_3)$
\Else
  \State Sample $b_s\sim\mathcal N(0,\sigma_{bs}^2)$
\EndIf
\State $T_0^p\gets T_0$
\For{$k=1,\ldots,N-1$}
  \State $\Delta T_k\gets T_{k-1}^{-1}T_k$
  \State Sample angular increments using
  Eq.~\ref{eq:noise_rotation}
  \State Construct $\delta R_k$ using the profile's rotation axes
  \If{time-dependent translation}
    \State Sample $\delta\mathbf t_k$ using
    Eq.~\ref{eq:noise_time_translation}
  \Else
    \State Sample $\delta\mathbf t_k$ using
    Eq.~\ref{eq:noise_distance_translation}
  \EndIf
  \State Compose $T_k^p$ using Eq.~\ref{eq:noise_composition}
\EndFor
\State \Return $\{T_k^p\}$
\end{algorithmic}
\end{algorithm}

\begin{algorithm}[t]
\caption{Independent pose-prior corruption}
\label{alg:iid}
\begin{algorithmic}[1]
\Require Camera-to-world poses $\{T_k\}$,
profile parameters, severity $g$
\State $\sigma_R\gets g\sigma_R$;
$\sigma_p\gets g\sigma_p$
\For{$k=0,\ldots,N-1$}
  \State Sample $\boldsymbol\epsilon_{R,k}
  \sim\mathcal N(\mathbf0,\sigma_R^2I_3)$
  \State Sample $\boldsymbol\epsilon_{p,k}
  \sim\mathcal N(\mathbf0,\sigma_p^2I_3)$
  \State Compute $T_k^p$ using Eq.~\ref{eq:noise_independent}
\EndFor
\State \Return $\{T_k^p\}$
\end{algorithmic}
\end{algorithm}

\paragraph{Evaluation corruption.}
For the pose-corruption conditions, evaluation generates a complete
corrupted trajectory for each scene and then extracts the selected
windows.
It does not load a fifth cached training realization.
The random generator is initialized from the evaluation seed,
a CRC32 key of the canonical scene path, and a fixed stream identifier.
The main evaluation uses a profile multiplier of 1, without additional
random severity sampling.
Time steps are derived from the declared nominal frame rate.
Camera intrinsics and depth receive no additional corruption.

\paragraph{Reproducibility.}
Corruption profiles, source-code versions, random seeds, and cached
training realizations are recorded separately.
Parameters describing the current implementation must match those
used to generate the training inputs.
Existing caches are not regenerated as part of result reporting.

\subsection{Visualization of Pose-Prior Corruptions}
\label{app:noise_visualization}

\begin{figure}[p]
\centering

\noisetrajpanel{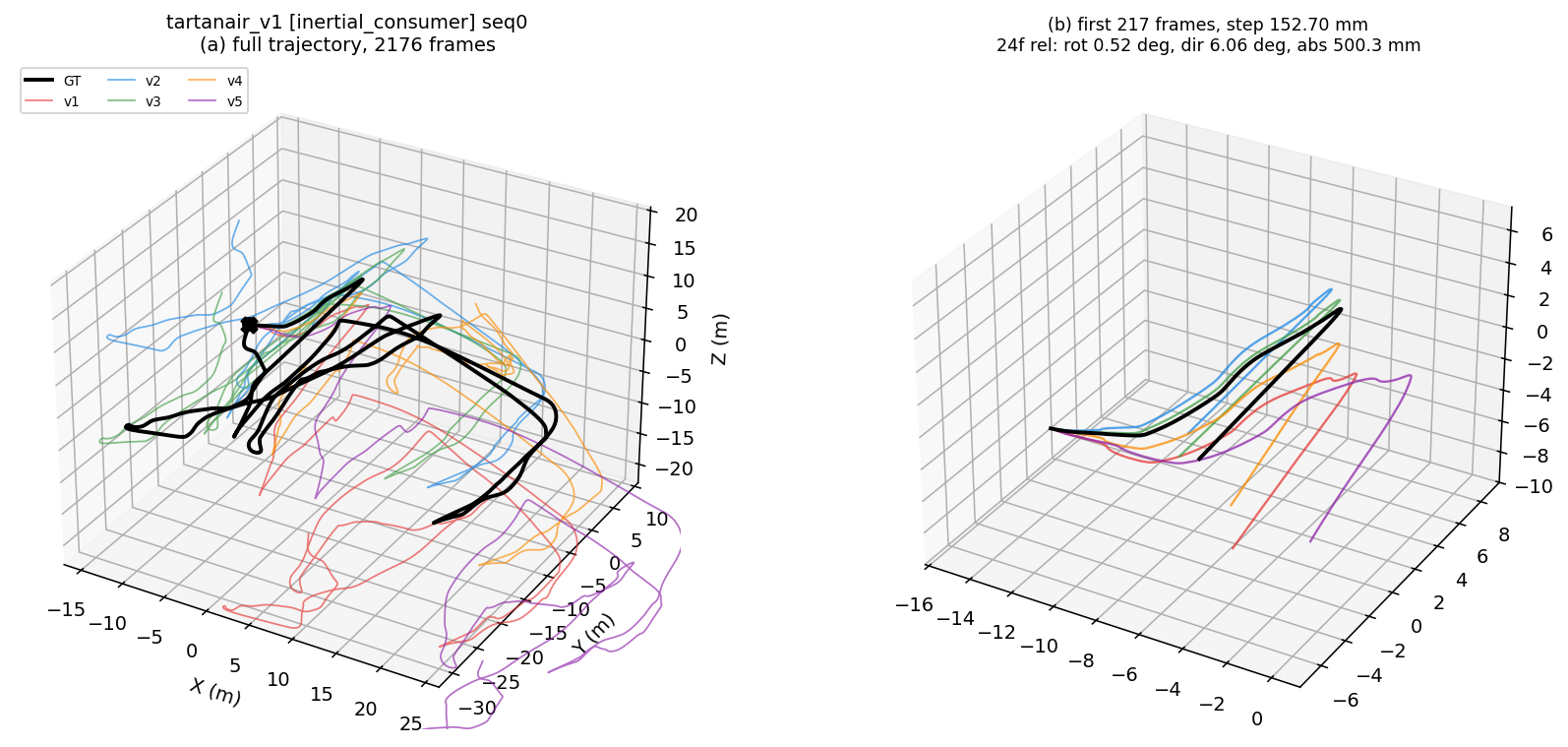}
{(a) TartanAir-V1: example 1}
\hfill
\noisetrajpanel{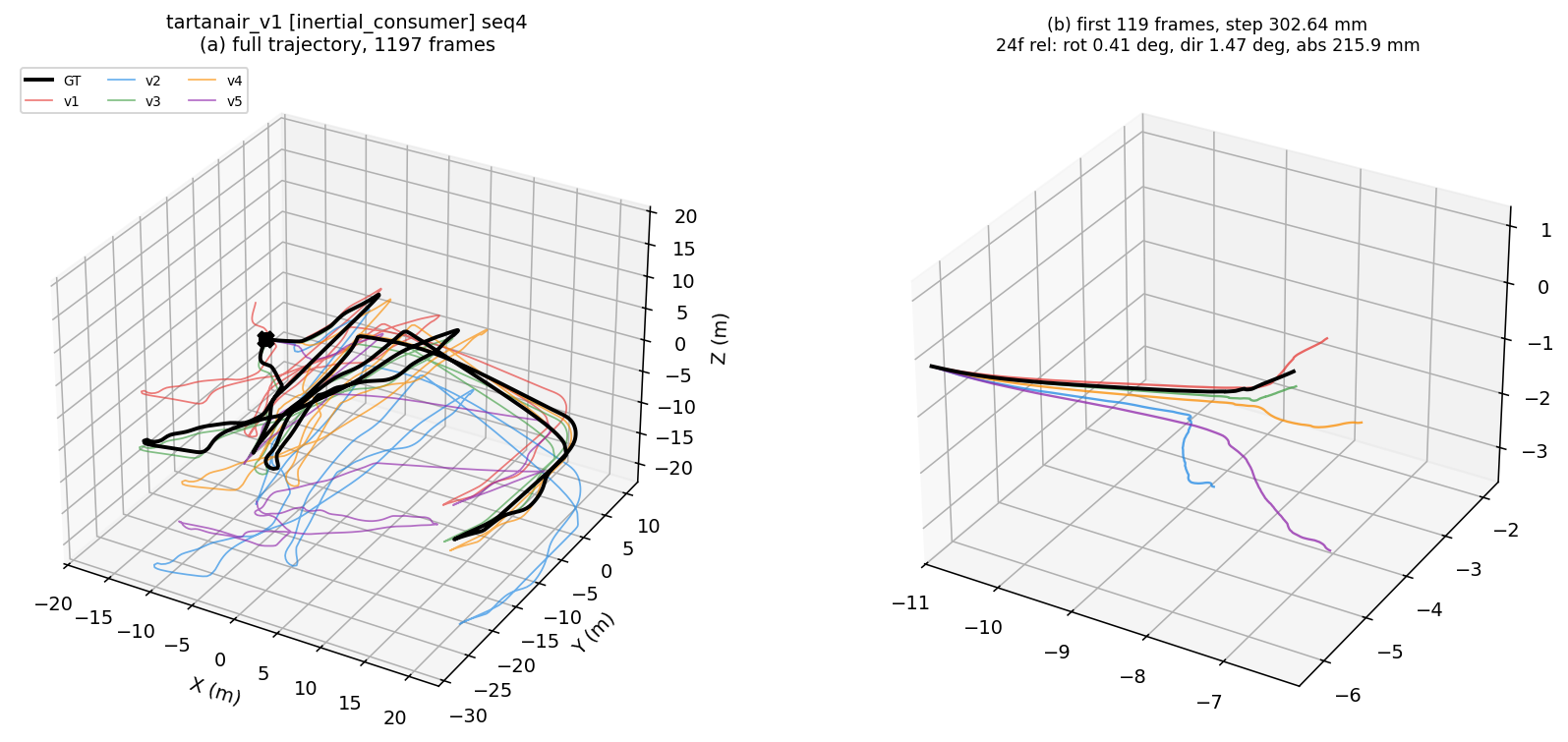}
{(b) TartanAir-V1: example 2}

\par\medskip

\noisetrajpanel{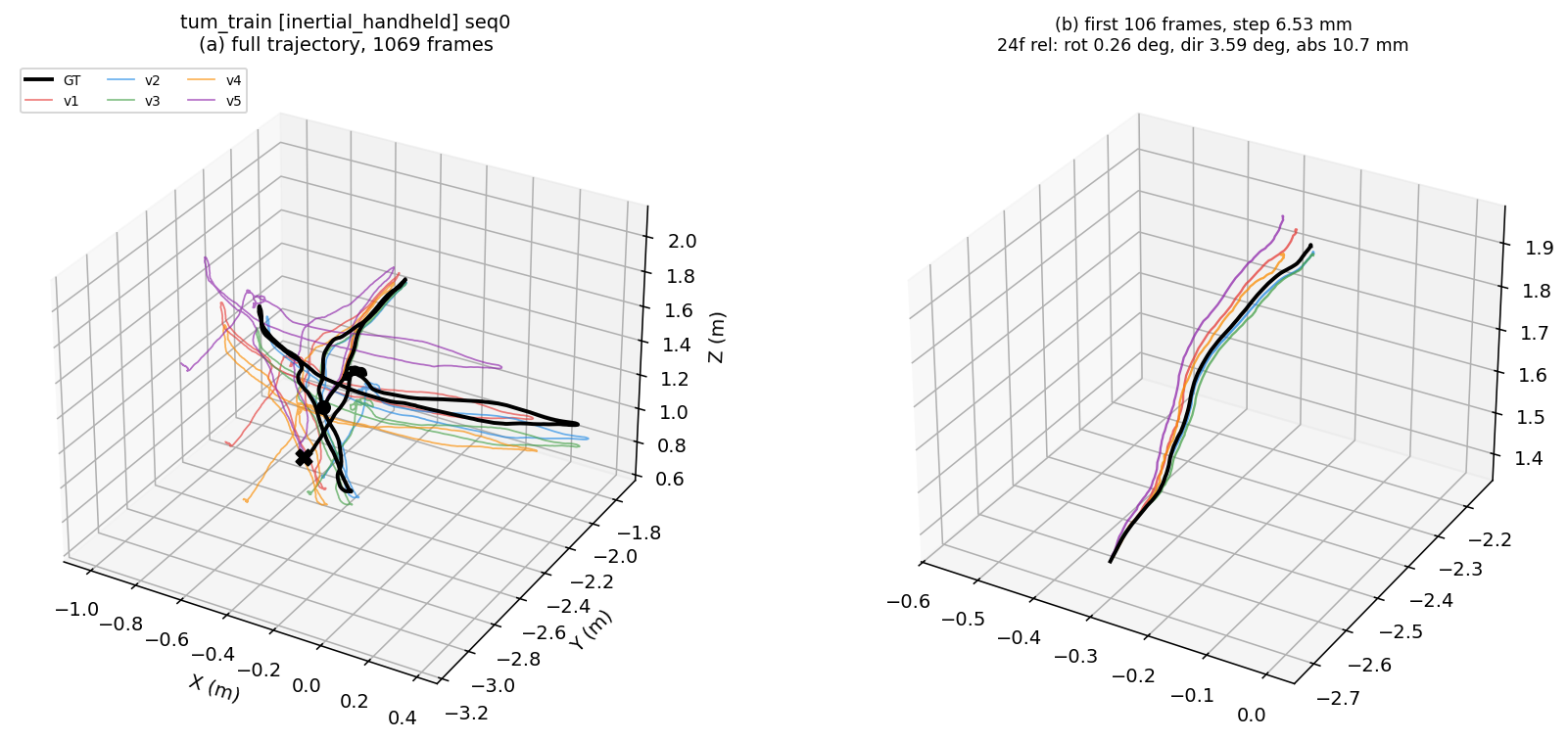}
{(c) TUM RGB-D: example 1}
\hfill
\noisetrajpanel{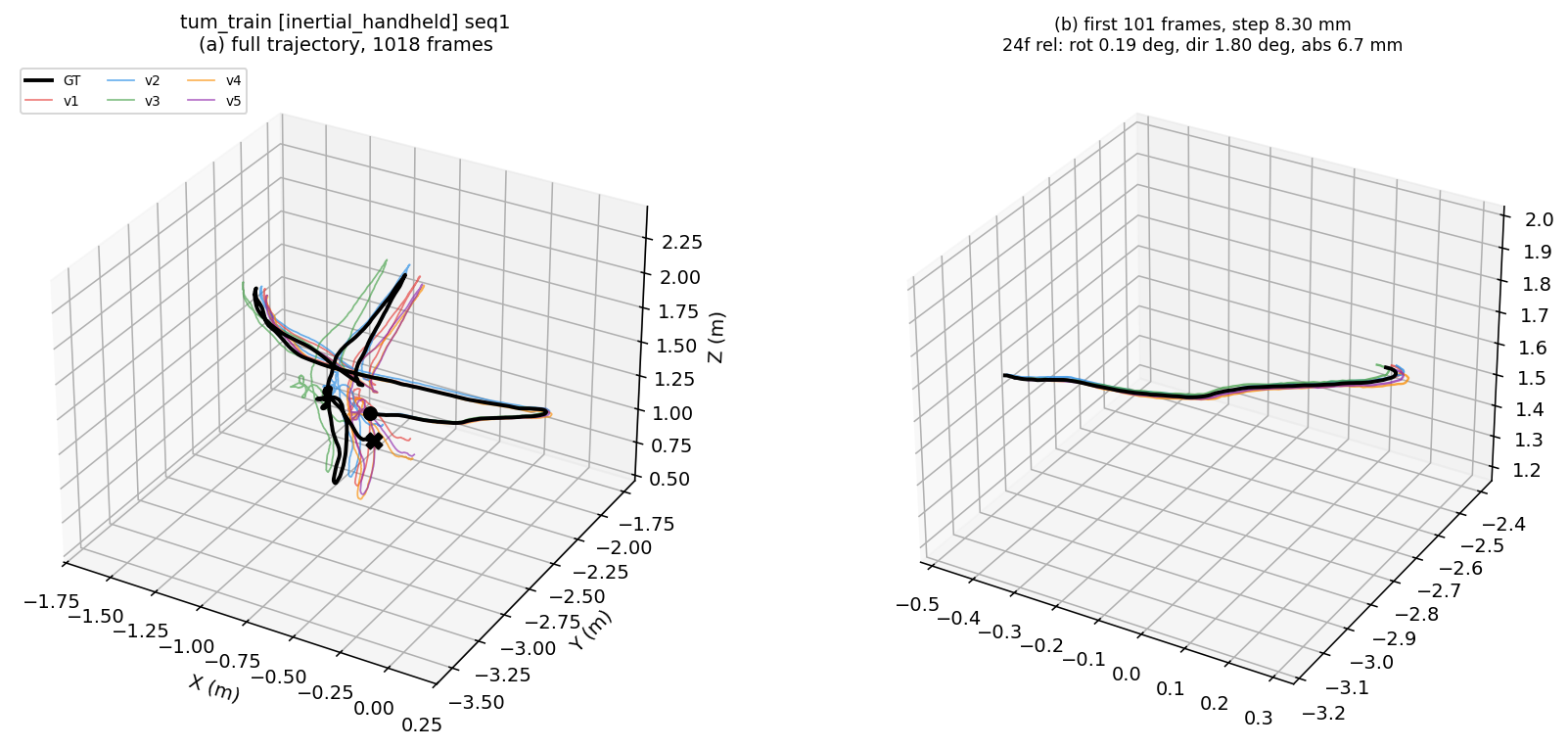}
{(d) TUM RGB-D: example 2}

\par\medskip

\noisetrajpanel{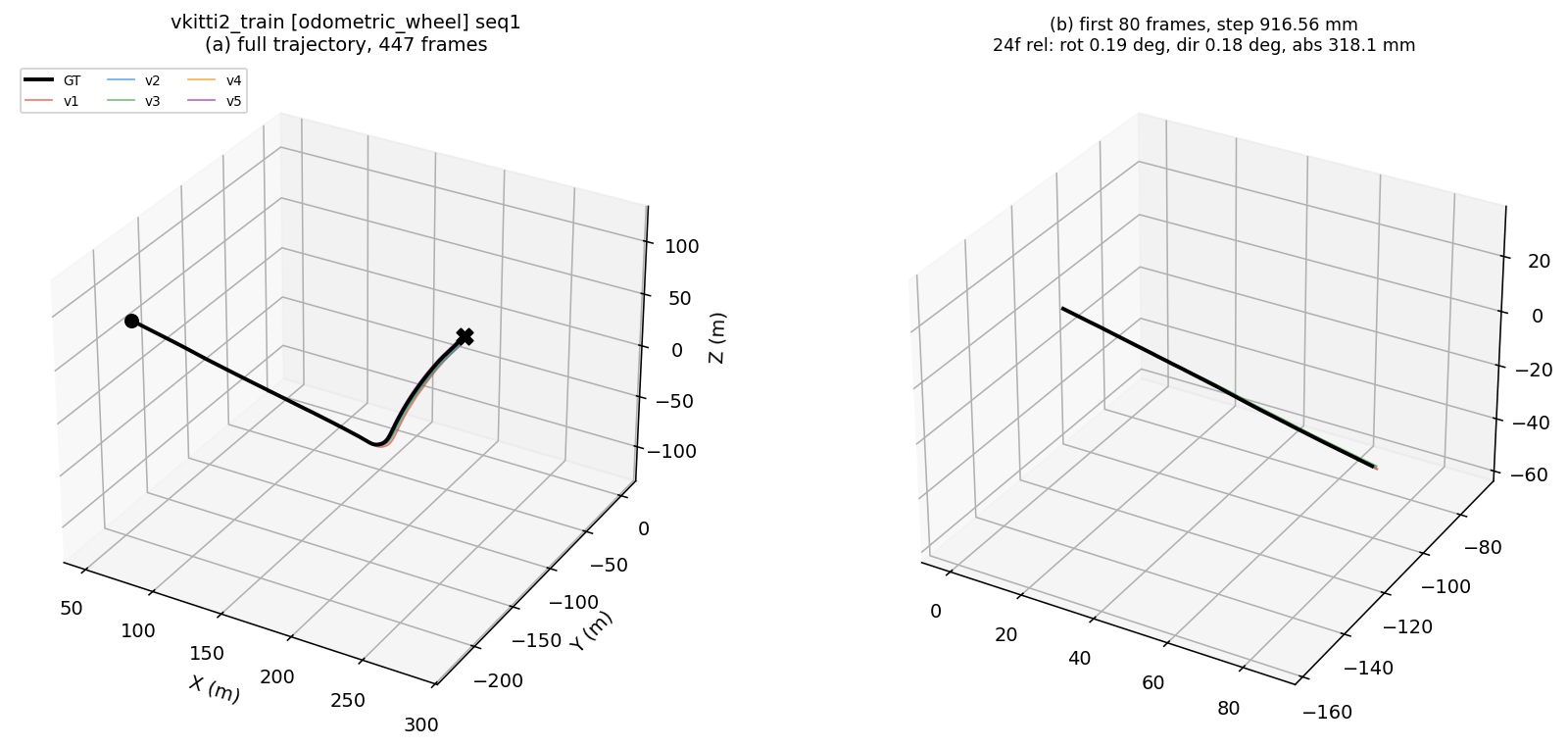}
{(e) Virtual KITTI 2: example 1}
\hfill
\noisetrajpanel{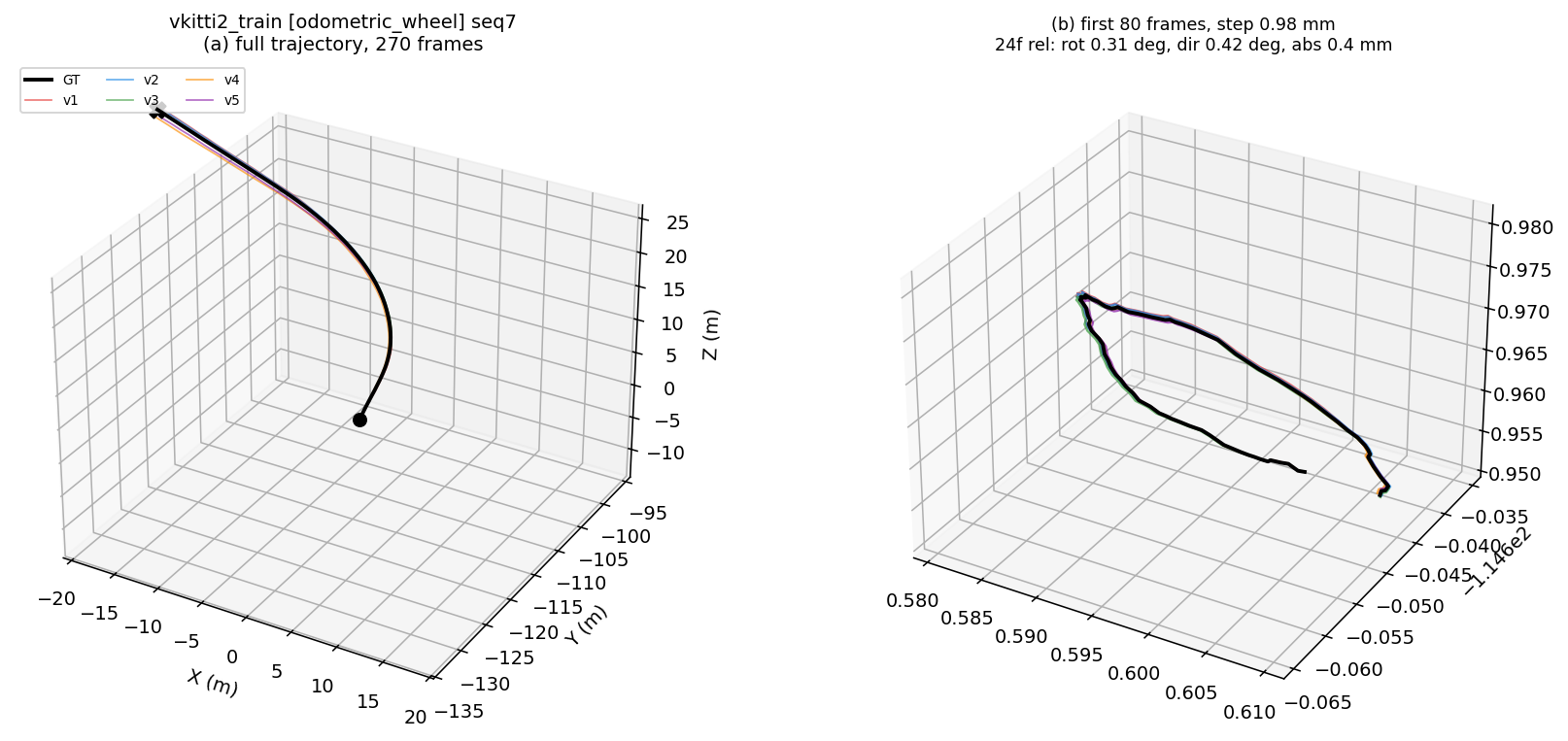}
{(f) Virtual KITTI 2: example 2}

\par\medskip

\noisetrajpanel{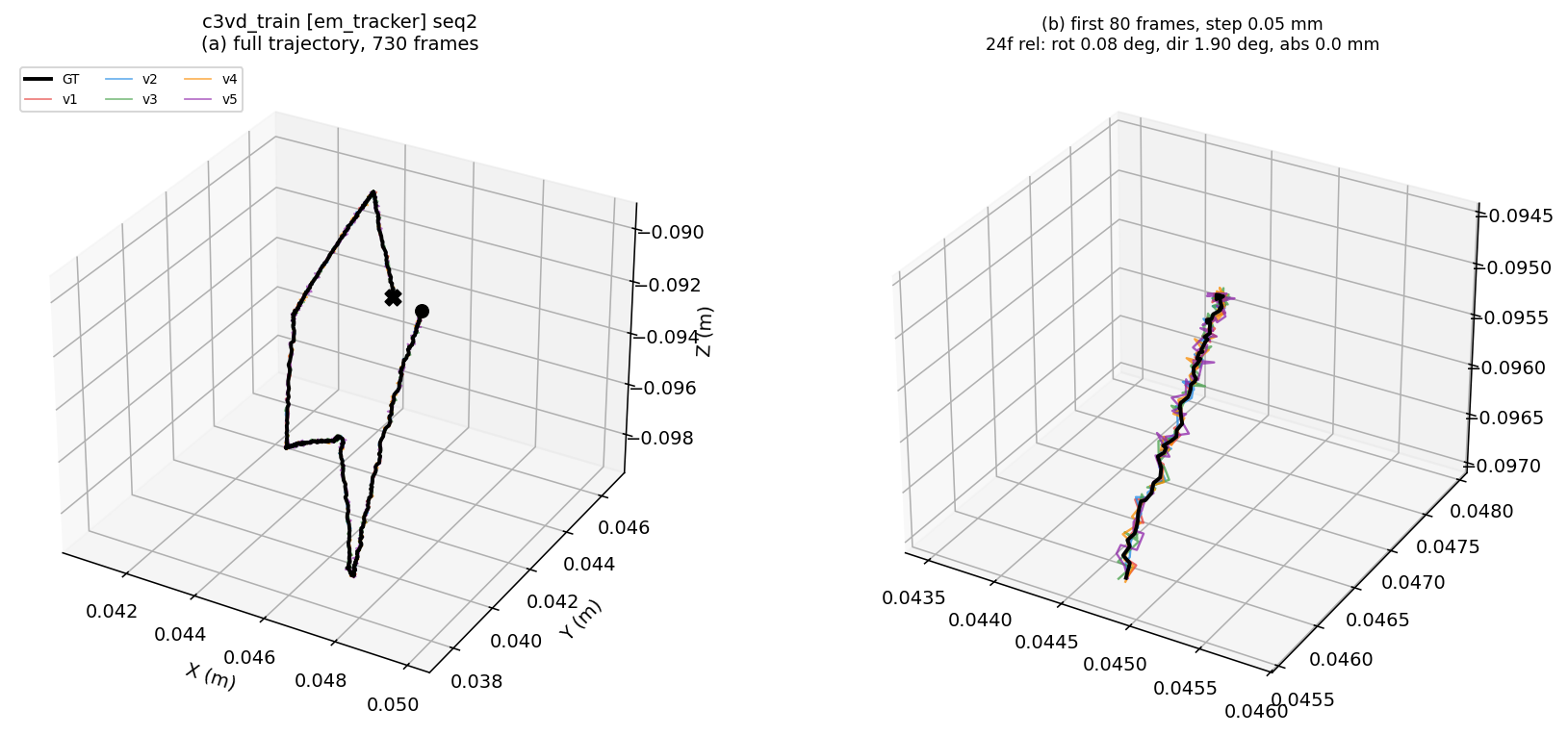}
{(g) C3VD: example 1}
\hfill
\noisetrajpanel{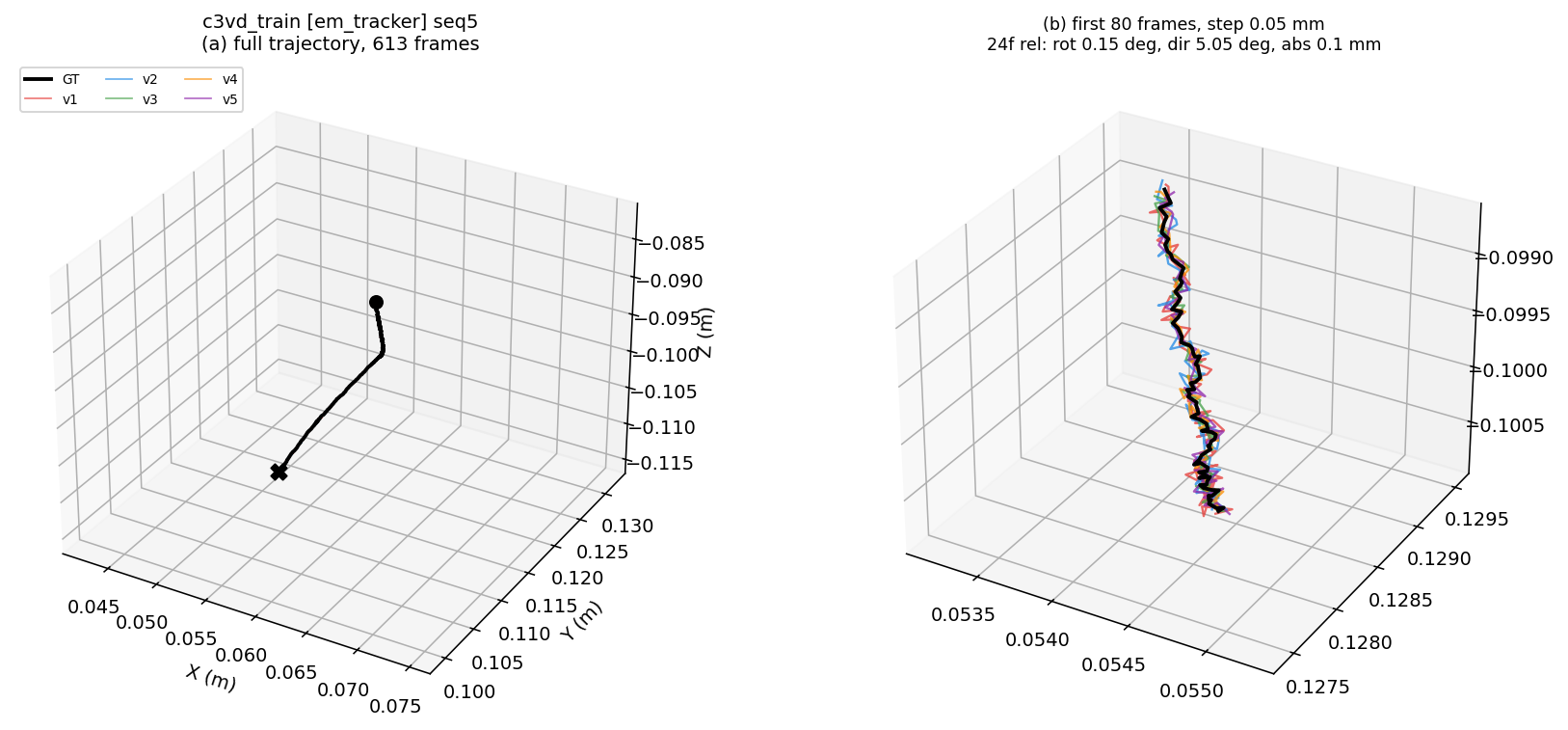}
{(h) C3VD: example 2}

\caption{\textbf{Pose-prior corruption in the training data.}
Two examples are shown for each dataset.
Within each panel, the left plot displays the complete trajectory,
while the right plot displays a selected segment of the same
ground-truth and corrupted trajectories.
Colors identify the ground truth and corruption realizations,
as indicated by the plot legends.
Ground-truth and corrupted poses share a coordinate system,
without independently fitted trajectory or scale alignment.
Spatial units and segment information are provided in the plots.}
\label{fig:noise_examples}
\end{figure}

Figure~\ref{fig:noise_examples} presents two training-trajectory
examples from each of the four datasets.
Within each panel, the left plot shows the complete trajectory,
and the right plot shows a selected segment from the same example.
The segment is extracted from the displayed corruption realizations,
rather than generated using new noise samples.
The global and local views illustrate the spatial structure of
accumulated drift and independent pose perturbations.

\subsection{Training and Checkpoint Details}
\label{app:training_details}

The matched prior-utilization comparison trains the released
OmniVGGT model with and without PRC for two epochs.
The final metric model is initialized from the trained PRC variant
and receives two additional epochs.
It therefore has a larger fine-tuning budget than either model
in the PRC ablation.

The scale stage uses four GPUs, 24 nominal image slots per GPU,
and three gradient-accumulation steps.
Learning rates are $10^{-5}$ for the backbone and geometric heads,
$5\times10^{-6}$ for the patch embedding, and
$2\times10^{-5}$ for the scale branch.
Training uses 500 warmup steps and cosine decay to 0.1 times
the initial learning rate.
The scale-loss weight is 1, and depth-based scale supervision
requires at least 100 valid pixels.

The data mixture contains 95,004 sampled training items per epoch.
This is a sampling budget, not a count of unique images.
Split manifests, checkpoint hashes, effective run configurations,
and evaluation-window identities are retained with the experiment
records.

\subsection{Evaluation Protocol}
\label{app:evaluation_protocol}

\paragraph{Windows and input selection.}
Evaluation uses ten-frame windows with a frame stride of three.
Each window spans 28 frames of the source sequence.
Up to ten windows are distributed across each eligible sequence,
without overlap between their temporal spans.
Sequences shorter than the required span are excluded according
to ground-truth availability.
Images are resized to width 518, with heights rounded to a
multiple of 14; intrinsics are adjusted accordingly.
For a coverage ratio $\rho$, priors are supplied to the first
$\operatorname{round}(10\rho)$ views.
Camera and depth coverage are controlled separately.
The camera input consists of intrinsics and poses together.

\paragraph{Relative-pose metrics.}
We evaluate every unordered image pair within a window.
Let $e_R$ denote relative-rotation error and $e_T^{\rm dir}$
the angle between the predicted and ground-truth relative
translation directions.
The sign-free error is
$e_T^{\rm sf}=\min(e_T^{\rm dir},180^\circ-e_T^{\rm dir})$.
We report RRA and RTA at $2^\circ$, $5^\circ$, and $15^\circ$,
using strict threshold comparisons.
Joint pose error is
$e_{\rm pose}=\max(e_R,e_T)$.
AUC@$15^\circ$ and AUC@$30^\circ$ use the cumulative mean of
the one-degree error histogram, matching the evaluator.
Unless marked ``dir'', RTA and AUC use sign-free translation errors.

Ground-truth baselines shorter than $10^{-6}$ in the processed
metric coordinates are excluded from translation and joint-pose
metrics, but remain eligible for rotation metrics.
An exactly zero predicted baseline against a valid ground-truth
baseline receives a translation error of $90^\circ$.
Non-finite pose predictions cause evaluation failure rather than
being removed from the scoring support.
Pose scores pool eligible pairs across evaluation windows.

\paragraph{Depth metrics.}
Depth validity is determined from ground truth.
Pixels must have finite depth greater than $10^{-4}$ and satisfy
the dataset-specific upper-depth limit, when one is configured.
Frames with fewer than 100 valid pixels are excluded.
For the scale-aligned results, each predicted depth map is multiplied
by the ratio of the ground-truth and predicted depth medians on
these pixels.
We then compute AbsRel and $\delta_1$, using the threshold 1.25.
The final scores average eligible frame-level metrics.
Predicted invalid depths on valid ground-truth pixels cause a
failure rather than a reduction in evaluation support.
We also report scores restricted to views without supplied depth.

\paragraph{Trajectory alignment.}
The ATE values accompanying the prior-utilization experiments use
Sim(3) alignment and measure trajectory shape up to scale.
They are not evidence of metric-scale accuracy.
Metric evaluation instead uses rigid SE(3) alignment, without
fitting a scale to test ground truth.

\paragraph{Metric reconstruction.}
Metric depth is evaluated on the same ground-truth-valid support as
aligned depth, but no median alignment is applied.
ATE-SE(3) is the root-mean-square camera-center error after fitting
rotation and translation only.
The reported value averages window-level errors.
C3VD trajectory errors are converted from meters to millimeters only
for display.

For the point-norm scale statistic, each ground-truth and predicted
reconstruction is expressed in its respective first-camera frame.
We average point distances over the common valid support and compute
their absolute relative difference.
VPGGT points are obtained from predicted Z-depth, predicted intrinsics,
and predicted poses.
MapAnything uses its predicted local geometry and camera poses.
This evaluation adopts the scale statistic, but not the datasets or
view-sampling protocol, of the original MapAnything benchmark.

\paragraph{Supplied-prior scale baseline.}
For the PRC-only scale baseline, let $\hat b_{ij}$ be a predicted
camera-center distance and $b^p_{ij}$ its supplied-prior counterpart.
A shared positive scale is obtained as
\begin{equation}
s_p =
\exp\left(
\operatorname{median}_{(i,j)\in\Omega_p}
[\log b^p_{ij}-\log\hat b_{ij}]
\right).
\end{equation}
The support uses nondegenerate supplied-camera baselines.
The fitted scale is applied to all translations and depth maps in the
window.
Evaluation ground truth is not used in this fit.

\paragraph{Implementation and aggregation.}
The external comparison reuses the data loading, window selection,
pose corruption, angular metrics, and aligned-depth functions of the
prior-utilization evaluator.
Additional metric scores are stored separately.
Pose metrics pool image-pair errors; depth metrics average eligible
frame scores; metric scale and trajectory metrics average window scores.
Prediction confidence does not alter scoring support.
Incomplete executions are not treated as completed benchmark rows.

\begin{table}[t]
\centering
\small
\caption{Dataset-specific evaluation settings.
Depth limits apply to ground-truth scoring support.
The PNG divisor is used only for encoded PNG depth.}
\label{tab:eval_ranges}
\begin{tabular}{@{}lrrc@{}}
\toprule
Dataset & Nominal FPS & Maximum depth (m) & PNG divisor\\
\midrule
TartanAir-V2 & 10 & 500 & --\\
Virtual KITTI 2 & 10 & 655 & 100\\
TUM RGB-D & 30 & 10 & --\\
C3VD & 30 & 0.15 & --\\
\bottomrule
\end{tabular}
\end{table}

\paragraph{Prior-retention support.}
AA, AU, and UU denote pairs with two, one, and zero supplied
camera priors, respectively.
The input prior is scored only on AA pairs.
At partial coverage, its scores are compared with the model's
AA scores, not with all-pair prediction scores.
The supplied prior is a reference, not an upper bound, when
camera poses are corrupted.

\paragraph{Gap and closure.}
For a higher-is-better score $M$, the signed prior gap is
$d_M=M_{\rm prior}-M_{\rm out}$.
A positive value indicates prior dilution.
When $M_{\rm prior}>M_0$, the closure is
$\kappa_M=(M_{\rm out}-M_0)/(M_{\rm prior}-M_0)$,
where $M_0$ uses the same model, windows, depth inputs, and
evaluation pairs without the camera bundle.
Because the camera bundle contains both intrinsics and poses,
this reference removes both.
Closure is undefined when the prior does not outperform the
reference.
For partial coverage, closure additionally requires reaggregating
the no-camera predictions on the same AA pairs.






\subsection{Complete Matched-Training Results}
\label{app:prc_full_results}

C1 denotes OmniVGGT-trained and C2 denotes VPGGT (PRC only).
Both are initialized from the same released checkpoint and trained
for two epochs with matched data, optimization settings, and
training budgets.
Their architectural difference is PRC.
Neither model contains the metric-scale branch.
Matched configurations do not imply identical realized stochastic
training samples at every optimization step.

At full camera-prior coverage without depth input, C2 improves
RTA@$5^\circ$ and AUC@$30^\circ$ over C1 on all four datasets
for both exact and corrupted poses.
For exact priors, the RTA gains are 3.33, 0.10, 15.00, and
33.58 percentage points on TartanAir-V2, Virtual KITTI 2,
TUM RGB-D, and C3VD, respectively.
The corresponding noisy-prior gains are 2.65, 0.13, 8.11,
and 12.84 points.

The improvements depend on input coverage and the evaluated metric.
At 30\% exact camera coverage on TartanAir-V2, RTA decreases
from 46.28 to 43.81.
At 30\% noisy camera coverage on C3VD, it decreases from
27.65 to 26.02.
Rotation accuracy does not improve uniformly.
These exceptions are retained in the complete tables.

Input-prior scores further distinguish improvement over a learned
baseline from improvement over the supplied observations.
With full noisy camera priors and no depth input, C2 exceeds
the TUM RGB-D prior in RTA@$5^\circ$ but not in AUC@$30^\circ$.
On C3VD, it exceeds the prior in AUC but not in RTA@$5^\circ$.
Consequently, prior retention and refinement are reported separately
for each metric.

\begin{table}[p]
\centering
\caption{\textbf{Complete matched-training results on TartanAir-V2.} C1: OmniVGGT-trained; C2: VPGGT (PRC only). Bold marks the better displayed score within each matched pair. Angular thresholds are in degrees; angular scores and $\delta_1$ are percentages.}
\label{tab:trained-v2-compact}
\begingroup\scriptsize
\setlength{\tabcolsep}{1.1pt}
\renewcommand{\arraystretch}{0.90}
\textbf{(a) Sign-free pose metrics}\par\smallskip
\begin{tabular*}{\linewidth}{@{\extracolsep{\fill}}rrclrrrrrrrr@{}}
\toprule
D & C & Pose & Model & R@2$\uparrow$ & R@5$\uparrow$ & R@15$\uparrow$ & T@2$\uparrow$ & T@5$\uparrow$ & T@15$\uparrow$ & A@15$\uparrow$ & A@30$\uparrow$ \\
\midrule
0 & 0 & -- & C1 & 58.52 & \textbf{88.69} & \textbf{96.46} & 10.49 & \textbf{46.48} & \textbf{88.31} & \textbf{58.07} & \textbf{75.01} \\
0 & 0 & -- & C2 & \textbf{58.97} & 88.37 & 96.12 & \textbf{10.69} & 46.35 & 87.78 & 57.77 & 74.63 \\
\addlinespace[0.5pt]
30 & 0 & -- & C1 & \textbf{60.92} & \textbf{88.91} & \textbf{96.28} & 9.30 & \textbf{42.18} & \textbf{86.72} & \textbf{55.78} & \textbf{73.39} \\
30 & 0 & -- & C2 & 60.56 & 88.48 & 96.24 & \textbf{9.32} & 41.98 & 86.39 & 55.50 & 73.12 \\
\addlinespace[0.5pt]
50 & 0 & -- & C1 & \textbf{62.41} & \textbf{89.00} & \textbf{96.03} & \textbf{9.04} & \textbf{41.86} & \textbf{86.26} & \textbf{55.49} & \textbf{73.01} \\
50 & 0 & -- & C2 & 62.32 & 88.78 & 95.96 & 8.99 & 41.66 & 85.77 & 55.12 & 72.77 \\
\addlinespace[0.5pt]
70 & 0 & -- & C1 & \textbf{64.55} & \textbf{89.48} & 96.00 & \textbf{8.87} & \textbf{41.28} & \textbf{86.35} & \textbf{55.42} & \textbf{72.90} \\
70 & 0 & -- & C2 & 64.30 & 89.34 & \textbf{96.11} & 8.48 & 40.72 & 85.92 & 55.07 & 72.67 \\
\addlinespace[0.5pt]
100 & 0 & -- & C1 & 67.02 & 90.17 & \textbf{96.12} & \textbf{8.33} & \textbf{40.87} & 86.62 & \textbf{55.53} & 73.01 \\
100 & 0 & -- & C2 & \textbf{67.26} & \textbf{90.32} & 96.03 & 8.14 & 40.40 & \textbf{86.64} & 55.34 & \textbf{73.04} \\
\addlinespace[0.5pt]
0 & 30 & GT & C1 & \textbf{62.22} & \textbf{89.37} & \textbf{96.46} & \textbf{12.13} & \textbf{46.28} & \textbf{84.95} & \textbf{55.86} & \textbf{72.91} \\
0 & 30 & GT & C2 & 62.05 & 89.13 & 96.28 & 11.46 & 43.81 & 82.52 & 53.62 & 70.92 \\
\addlinespace[0.5pt]
0 & 50 & GT & C1 & \textbf{64.34} & \textbf{89.81} & \textbf{96.61} & \textbf{13.86} & \textbf{49.60} & \textbf{85.73} & \textbf{57.51} & \textbf{74.03} \\
0 & 50 & GT & C2 & 64.31 & 89.48 & 96.59 & 13.72 & 46.96 & 83.26 & 55.14 & 72.00 \\
\addlinespace[0.5pt]
0 & 70 & GT & C1 & 69.40 & \textbf{91.25} & \textbf{97.14} & 17.68 & \textbf{57.03} & \textbf{87.90} & \textbf{61.93} & \textbf{76.87} \\
0 & 70 & GT & C2 & \textbf{70.03} & 91.03 & 97.02 & \textbf{18.79} & 55.55 & 86.12 & 60.54 & 75.65 \\
\addlinespace[0.5pt]
0 & 100 & GT & C1 & 85.50 & 95.83 & 98.36 & 25.89 & 73.32 & 94.21 & 73.09 & 84.67 \\
0 & 100 & GT & C2 & \textbf{88.55} & \textbf{95.85} & \textbf{98.49} & \textbf{31.41} & \textbf{76.65} & \textbf{94.34} & \textbf{75.33} & \textbf{85.85} \\
\addlinespace[0.5pt]
0 & 30 & N & C1 & \textbf{62.22} & \textbf{89.37} & \textbf{96.46} & \textbf{12.11} & \textbf{46.31} & \textbf{84.94} & \textbf{55.83} & \textbf{72.90} \\
0 & 30 & N & C2 & 62.10 & 89.11 & 96.29 & 11.49 & 43.68 & 82.54 & 53.57 & 70.89 \\
\addlinespace[0.5pt]
0 & 50 & N & C1 & \textbf{64.27} & \textbf{89.81} & \textbf{96.61} & \textbf{13.87} & \textbf{49.63} & \textbf{85.71} & \textbf{57.49} & \textbf{74.02} \\
0 & 50 & N & C2 & 64.18 & 89.47 & 96.59 & 13.55 & 46.83 & 83.24 & 55.04 & 71.94 \\
\addlinespace[0.5pt]
0 & 70 & N & C1 & 69.21 & \textbf{91.26} & \textbf{97.14} & 17.65 & \textbf{57.08} & \textbf{87.91} & \textbf{61.91} & \textbf{76.86} \\
0 & 70 & N & C2 & \textbf{69.82} & 91.02 & 97.01 & \textbf{18.19} & 55.28 & 86.03 & 60.32 & 75.52 \\
\addlinespace[0.5pt]
0 & 100 & N & C1 & 85.37 & \textbf{95.82} & 98.34 & 25.94 & 73.12 & 94.20 & 73.06 & 84.65 \\
0 & 100 & N & C2 & \textbf{87.89} & 95.81 & \textbf{98.47} & \textbf{29.32} & \textbf{75.77} & \textbf{94.24} & \textbf{74.65} & \textbf{85.50} \\
\addlinespace[0.5pt]
100 & 100 & GT & C1 & 87.33 & 95.87 & 98.13 & 24.08 & 74.66 & 94.06 & 73.36 & 84.76 \\
100 & 100 & GT & C2 & \textbf{90.59} & \textbf{96.42} & \textbf{98.68} & \textbf{27.59} & \textbf{76.87} & \textbf{94.60} & \textbf{74.99} & \textbf{85.80} \\
\addlinespace[0.5pt]
100 & 100 & N & C1 & 87.06 & 95.86 & 98.14 & 23.86 & 74.65 & 94.05 & 73.31 & 84.72 \\
100 & 100 & N & C2 & \textbf{89.89} & \textbf{96.39} & \textbf{98.70} & \textbf{25.99} & \textbf{76.05} & \textbf{94.56} & \textbf{74.44} & \textbf{85.51} \\
\addlinespace[0.5pt]
\bottomrule
\end{tabular*}
\par\medskip
\begin{minipage}[t]{0.49\linewidth}
\centering
\textbf{(b) Directed translation and joint pose}\par\smallskip
\begin{tabular*}{\linewidth}{@{\extracolsep{\fill}}rrclrrrrr@{}}
\toprule
D & C & Pose & Model & T@2$\uparrow$ & T@5$\uparrow$ & T@15$\uparrow$ & A@15$\uparrow$ & A@30$\uparrow$ \\
\midrule
0 & 0 & -- & C1 & 10.48 & \textbf{46.43} & \textbf{87.99} & \textbf{57.97} & \textbf{74.81} \\
0 & 0 & -- & C2 & \textbf{10.69} & 46.30 & 87.46 & 57.68 & 74.41 \\
\addlinespace[0.5pt]
30 & 0 & -- & C1 & 9.29 & \textbf{42.11} & \textbf{86.41} & \textbf{55.68} & \textbf{73.15} \\
30 & 0 & -- & C2 & \textbf{9.31} & 41.92 & 86.08 & 55.39 & 72.88 \\
\addlinespace[0.5pt]
50 & 0 & -- & C1 & \textbf{9.03} & \textbf{41.77} & \textbf{85.92} & \textbf{55.38} & \textbf{72.78} \\
50 & 0 & -- & C2 & 8.98 & 41.56 & 85.44 & 55.01 & 72.49 \\
\addlinespace[0.5pt]
70 & 0 & -- & C1 & \textbf{8.84} & \textbf{41.17} & \textbf{85.88} & \textbf{55.29} & \textbf{72.60} \\
70 & 0 & -- & C2 & 8.47 & 40.65 & 85.55 & 54.93 & 72.37 \\
\addlinespace[0.5pt]
100 & 0 & -- & C1 & \textbf{8.32} & \textbf{40.77} & 86.17 & \textbf{55.39} & 72.73 \\
100 & 0 & -- & C2 & 8.13 & 40.33 & \textbf{86.29} & 55.21 & \textbf{72.77} \\
\addlinespace[0.5pt]
0 & 30 & GT & C1 & \textbf{12.13} & \textbf{46.23} & \textbf{84.64} & \textbf{55.75} & \textbf{72.71} \\
0 & 30 & GT & C2 & 11.45 & 43.76 & 82.20 & 53.51 & 70.68 \\
\addlinespace[0.5pt]
0 & 50 & GT & C1 & \textbf{13.85} & \textbf{49.54} & \textbf{85.42} & \textbf{57.41} & \textbf{73.84} \\
0 & 50 & GT & C2 & 13.72 & 46.92 & 82.97 & 55.03 & 71.78 \\
\addlinespace[0.5pt]
0 & 70 & GT & C1 & 17.68 & \textbf{57.00} & \textbf{87.62} & \textbf{61.84} & \textbf{76.69} \\
0 & 70 & GT & C2 & \textbf{18.78} & 55.51 & 85.86 & 60.46 & 75.46 \\
\addlinespace[0.5pt]
0 & 100 & GT & C1 & 25.89 & 73.32 & 94.18 & 73.08 & 84.65 \\
0 & 100 & GT & C2 & \textbf{31.41} & \textbf{76.65} & \textbf{94.32} & \textbf{75.32} & \textbf{85.82} \\
\addlinespace[0.5pt]
0 & 30 & N & C1 & \textbf{12.11} & \textbf{46.26} & \textbf{84.65} & \textbf{55.73} & \textbf{72.70} \\
0 & 30 & N & C2 & 11.48 & 43.63 & 82.22 & 53.47 & 70.65 \\
\addlinespace[0.5pt]
0 & 50 & N & C1 & \textbf{13.86} & \textbf{49.57} & \textbf{85.40} & \textbf{57.39} & \textbf{73.82} \\
0 & 50 & N & C2 & 13.55 & 46.79 & 82.94 & 54.93 & 71.72 \\
\addlinespace[0.5pt]
0 & 70 & N & C1 & 17.65 & \textbf{57.05} & \textbf{87.63} & \textbf{61.82} & \textbf{76.67} \\
0 & 70 & N & C2 & \textbf{18.18} & 55.25 & 85.76 & 60.24 & 75.33 \\
\addlinespace[0.5pt]
0 & 100 & N & C1 & 25.94 & 73.12 & 94.17 & 73.06 & 84.64 \\
0 & 100 & N & C2 & \textbf{29.32} & \textbf{75.76} & \textbf{94.21} & \textbf{74.64} & \textbf{85.47} \\
\addlinespace[0.5pt]
100 & 100 & GT & C1 & 24.08 & 74.66 & 93.99 & 73.33 & 84.70 \\
100 & 100 & GT & C2 & \textbf{27.59} & \textbf{76.87} & \textbf{94.58} & \textbf{74.98} & \textbf{85.76} \\
\addlinespace[0.5pt]
100 & 100 & N & C1 & 23.86 & 74.65 & 93.98 & 73.29 & 84.66 \\
100 & 100 & N & C2 & \textbf{25.99} & \textbf{76.05} & \textbf{94.53} & \textbf{74.43} & \textbf{85.47} \\
\addlinespace[0.5pt]
\bottomrule
\end{tabular*}
\end{minipage}\hfill
\begin{minipage}[t]{0.49\linewidth}
\centering
\textbf{(c) Per-frame median-aligned depth}\par\smallskip
\begin{tabular*}{\linewidth}{@{\extracolsep{\fill}}rrclrrrr@{}}
\toprule
D & C & Pose & Model & AbsRel$\downarrow$ & $\delta_1\uparrow$ & AbsRel-U$\downarrow$ & $\delta_{1,U}\uparrow$ \\
\midrule
0 & 0 & -- & C1 & 0.109 & \textbf{89.62} & 0.109 & \textbf{89.62} \\
0 & 0 & -- & C2 & 0.109 & 89.59 & 0.109 & 89.59 \\
\addlinespace[0.5pt]
30 & 0 & -- & C1 & 0.065 & \textbf{94.00} & \textbf{0.084} & \textbf{91.86} \\
30 & 0 & -- & C2 & 0.065 & 93.95 & 0.085 & 91.78 \\
\addlinespace[0.5pt]
50 & 0 & -- & C1 & 0.050 & \textbf{95.58} & \textbf{0.080} & \textbf{92.15} \\
50 & 0 & -- & C2 & 0.050 & 95.56 & 0.081 & 92.10 \\
\addlinespace[0.5pt]
70 & 0 & -- & C1 & 0.037 & \textbf{97.14} & 0.075 & \textbf{92.77} \\
70 & 0 & -- & C2 & 0.037 & 97.12 & 0.075 & 92.69 \\
\addlinespace[0.5pt]
100 & 0 & -- & C1 & 0.021 & 99.00 & -- & -- \\
100 & 0 & -- & C2 & 0.021 & \textbf{99.01} & -- & -- \\
\addlinespace[0.5pt]
0 & 30 & GT & C1 & \textbf{0.108} & \textbf{89.70} & \textbf{0.108} & \textbf{89.70} \\
0 & 30 & GT & C2 & 0.109 & 89.65 & 0.109 & 89.65 \\
\addlinespace[0.5pt]
0 & 50 & GT & C1 & 0.108 & \textbf{89.75} & 0.108 & \textbf{89.75} \\
0 & 50 & GT & C2 & 0.108 & 89.70 & 0.108 & 89.70 \\
\addlinespace[0.5pt]
0 & 70 & GT & C1 & \textbf{0.107} & \textbf{89.79} & \textbf{0.107} & \textbf{89.79} \\
0 & 70 & GT & C2 & 0.108 & 89.75 & 0.108 & 89.75 \\
\addlinespace[0.5pt]
0 & 100 & GT & C1 & 0.107 & \textbf{89.88} & 0.107 & \textbf{89.88} \\
0 & 100 & GT & C2 & 0.107 & 89.83 & 0.107 & 89.83 \\
\addlinespace[0.5pt]
0 & 30 & N & C1 & \textbf{0.108} & \textbf{89.70} & \textbf{0.108} & \textbf{89.70} \\
0 & 30 & N & C2 & 0.109 & 89.65 & 0.109 & 89.65 \\
\addlinespace[0.5pt]
0 & 50 & N & C1 & 0.108 & \textbf{89.75} & 0.108 & \textbf{89.75} \\
0 & 50 & N & C2 & 0.108 & 89.71 & 0.108 & 89.71 \\
\addlinespace[0.5pt]
0 & 70 & N & C1 & \textbf{0.107} & \textbf{89.79} & \textbf{0.107} & \textbf{89.79} \\
0 & 70 & N & C2 & 0.108 & 89.75 & 0.108 & 89.75 \\
\addlinespace[0.5pt]
0 & 100 & N & C1 & 0.107 & \textbf{89.88} & 0.107 & \textbf{89.88} \\
0 & 100 & N & C2 & 0.107 & 89.83 & 0.107 & 89.83 \\
\addlinespace[0.5pt]
100 & 100 & GT & C1 & 0.020 & 99.01 & -- & -- \\
100 & 100 & GT & C2 & 0.020 & 99.01 & -- & -- \\
\addlinespace[0.5pt]
100 & 100 & N & C1 & 0.020 & 99.01 & -- & -- \\
100 & 100 & N & C2 & 0.020 & 99.01 & -- & -- \\
\addlinespace[0.5pt]
\bottomrule
\end{tabular*}
\end{minipage}
\par\smallskip
R/T/A denote RRA/RTA/AUC; panel (b) uses directed errors. U denotes unprovided depth views. D/C are provided-view percentages.
\endgroup
\end{table}

\begin{table}[p]
\centering
\caption{\textbf{Complete matched-training results on Virtual KITTI 2.} C1: OmniVGGT-trained; C2: VPGGT (PRC only). Bold marks the better displayed score within each matched pair. Angular thresholds are in degrees; angular scores and $\delta_1$ are percentages.}
\label{tab:trained-vkitti-eval-compact}
\begingroup\scriptsize
\setlength{\tabcolsep}{1.1pt}
\renewcommand{\arraystretch}{0.90}
\textbf{(a) Sign-free pose metrics}\par\smallskip
\begin{tabular*}{\linewidth}{@{\extracolsep{\fill}}rrclrrrrrrrr@{}}
\toprule
D & C & Pose & Model & R@2$\uparrow$ & R@5$\uparrow$ & R@15$\uparrow$ & T@2$\uparrow$ & T@5$\uparrow$ & T@15$\uparrow$ & A@15$\uparrow$ & A@30$\uparrow$ \\
\midrule
0 & 0 & -- & C1 & 100.00 & 100.00 & 100.00 & \textbf{97.87} & 99.65 & 99.81 & \textbf{98.54} & \textbf{99.21} \\
0 & 0 & -- & C2 & 100.00 & 100.00 & 100.00 & 96.98 & \textbf{99.68} & \textbf{99.84} & 98.43 & 99.15 \\
\addlinespace[0.5pt]
30 & 0 & -- & C1 & 100.00 & 100.00 & 100.00 & \textbf{96.92} & 99.65 & 99.84 & \textbf{98.40} & \textbf{99.14} \\
30 & 0 & -- & C2 & 100.00 & 100.00 & 100.00 & 95.46 & \textbf{99.68} & 99.84 & 98.17 & 99.01 \\
\addlinespace[0.5pt]
50 & 0 & -- & C1 & \textbf{99.43} & 100.00 & 100.00 & \textbf{95.75} & \textbf{99.40} & \textbf{99.75} & \textbf{97.97} & \textbf{98.89} \\
50 & 0 & -- & C2 & 98.89 & 100.00 & 100.00 & 93.97 & 99.14 & 99.65 & 97.46 & 98.57 \\
\addlinespace[0.5pt]
70 & 0 & -- & C1 & 99.30 & 100.00 & 100.00 & \textbf{95.84} & \textbf{99.08} & \textbf{99.65} & \textbf{97.81} & \textbf{98.78} \\
70 & 0 & -- & C2 & \textbf{99.33} & 100.00 & 100.00 & 94.29 & 98.98 & 99.62 & 97.51 & 98.63 \\
\addlinespace[0.5pt]
100 & 0 & -- & C1 & \textbf{98.95} & 100.00 & 100.00 & \textbf{96.16} & 98.54 & 99.24 & \textbf{97.38} & 98.44 \\
100 & 0 & -- & C2 & 98.89 & 100.00 & 100.00 & 94.57 & \textbf{98.79} & \textbf{99.43} & 97.35 & \textbf{98.47} \\
\addlinespace[0.5pt]
0 & 30 & GT & C1 & 100.00 & 100.00 & 100.00 & \textbf{99.05} & 99.65 & \textbf{99.81} & \textbf{99.02} & \textbf{99.45} \\
0 & 30 & GT & C2 & 100.00 & 100.00 & 100.00 & 98.10 & \textbf{99.68} & 99.78 & 98.86 & 99.33 \\
\addlinespace[0.5pt]
0 & 50 & GT & C1 & 100.00 & 100.00 & 100.00 & \textbf{98.95} & \textbf{99.68} & \textbf{99.81} & \textbf{98.97} & \textbf{99.43} \\
0 & 50 & GT & C2 & 100.00 & 100.00 & 100.00 & 97.75 & 99.56 & 99.78 & 98.74 & 99.27 \\
\addlinespace[0.5pt]
0 & 70 & GT & C1 & 100.00 & 100.00 & 100.00 & \textbf{97.68} & \textbf{99.33} & \textbf{99.84} & \textbf{98.75} & \textbf{99.32} \\
0 & 70 & GT & C2 & 100.00 & 100.00 & 100.00 & 97.14 & 99.02 & 99.56 & 98.35 & 99.05 \\
\addlinespace[0.5pt]
0 & 100 & GT & C1 & 100.00 & 100.00 & 100.00 & 99.27 & 99.71 & 99.84 & 99.40 & 99.66 \\
0 & 100 & GT & C2 & 100.00 & 100.00 & 100.00 & \textbf{99.59} & \textbf{99.81} & \textbf{99.94} & \textbf{99.76} & \textbf{99.87} \\
\addlinespace[0.5pt]
0 & 30 & N & C1 & 100.00 & 100.00 & 100.00 & \textbf{99.08} & 99.65 & \textbf{99.81} & \textbf{99.02} & \textbf{99.45} \\
0 & 30 & N & C2 & 100.00 & 100.00 & 100.00 & 98.06 & \textbf{99.68} & 99.78 & 98.85 & 99.33 \\
\addlinespace[0.5pt]
0 & 50 & N & C1 & 100.00 & 100.00 & 100.00 & \textbf{98.92} & \textbf{99.68} & \textbf{99.81} & \textbf{98.98} & \textbf{99.43} \\
0 & 50 & N & C2 & 100.00 & 100.00 & 100.00 & 97.78 & 99.56 & 99.78 & 98.74 & 99.27 \\
\addlinespace[0.5pt]
0 & 70 & N & C1 & 100.00 & 100.00 & 100.00 & \textbf{97.59} & \textbf{99.27} & \textbf{99.84} & \textbf{98.74} & \textbf{99.31} \\
0 & 70 & N & C2 & 100.00 & 100.00 & 100.00 & 97.21 & 99.02 & 99.56 & 98.35 & 99.05 \\
\addlinespace[0.5pt]
0 & 100 & N & C1 & 100.00 & 100.00 & 100.00 & 99.24 & 99.68 & 99.84 & 99.39 & 99.66 \\
0 & 100 & N & C2 & 100.00 & 100.00 & 100.00 & \textbf{99.59} & \textbf{99.81} & \textbf{99.94} & \textbf{99.72} & \textbf{99.85} \\
\addlinespace[0.5pt]
100 & 100 & GT & C1 & 100.00 & 100.00 & 100.00 & 98.73 & 99.75 & 99.87 & 99.17 & 99.54 \\
100 & 100 & GT & C2 & 100.00 & 100.00 & 100.00 & \textbf{99.27} & \textbf{99.81} & \textbf{99.94} & \textbf{99.64} & \textbf{99.79} \\
\addlinespace[0.5pt]
100 & 100 & N & C1 & 100.00 & 100.00 & 100.00 & 98.63 & 99.75 & 99.87 & 99.13 & 99.52 \\
100 & 100 & N & C2 & 100.00 & 100.00 & 100.00 & \textbf{99.27} & \textbf{99.78} & \textbf{99.94} & \textbf{99.60} & \textbf{99.77} \\
\addlinespace[0.5pt]
\bottomrule
\end{tabular*}
\par\medskip
\begin{minipage}[t]{0.49\linewidth}
\centering
\textbf{(b) Directed translation and joint pose}\par\smallskip
\begin{tabular*}{\linewidth}{@{\extracolsep{\fill}}rrclrrrrr@{}}
\toprule
D & C & Pose & Model & T@2$\uparrow$ & T@5$\uparrow$ & T@15$\uparrow$ & A@15$\uparrow$ & A@30$\uparrow$ \\
\midrule
0 & 0 & -- & C1 & \textbf{97.87} & 99.65 & 99.81 & \textbf{98.54} & \textbf{99.21} \\
0 & 0 & -- & C2 & 96.98 & \textbf{99.68} & \textbf{99.84} & 98.43 & 99.15 \\
\addlinespace[0.5pt]
30 & 0 & -- & C1 & \textbf{96.92} & 99.65 & 99.84 & \textbf{98.40} & \textbf{99.14} \\
30 & 0 & -- & C2 & 95.46 & \textbf{99.68} & 99.84 & 98.17 & 99.01 \\
\addlinespace[0.5pt]
50 & 0 & -- & C1 & \textbf{95.75} & \textbf{99.40} & \textbf{99.75} & \textbf{97.97} & \textbf{98.88} \\
50 & 0 & -- & C2 & 93.97 & 99.14 & 99.59 & 97.42 & 98.52 \\
\addlinespace[0.5pt]
70 & 0 & -- & C1 & \textbf{95.84} & \textbf{99.08} & \textbf{99.62} & \textbf{97.80} & \textbf{98.75} \\
70 & 0 & -- & C2 & 94.29 & 98.98 & 99.59 & 97.50 & 98.61 \\
\addlinespace[0.5pt]
100 & 0 & -- & C1 & \textbf{95.75} & \textbf{97.90} & \textbf{98.54} & \textbf{96.83} & \textbf{97.75} \\
100 & 0 & -- & C2 & 94.22 & 97.71 & 98.25 & 96.39 & 97.41 \\
\addlinespace[0.5pt]
0 & 30 & GT & C1 & \textbf{99.05} & 99.65 & \textbf{99.81} & \textbf{99.02} & \textbf{99.45} \\
0 & 30 & GT & C2 & 98.10 & \textbf{99.68} & 99.78 & 98.86 & 99.33 \\
\addlinespace[0.5pt]
0 & 50 & GT & C1 & \textbf{98.95} & \textbf{99.68} & \textbf{99.81} & \textbf{98.97} & \textbf{99.43} \\
0 & 50 & GT & C2 & 97.75 & 99.56 & 99.78 & 98.74 & 99.27 \\
\addlinespace[0.5pt]
0 & 70 & GT & C1 & \textbf{97.68} & \textbf{99.33} & \textbf{99.84} & \textbf{98.75} & \textbf{99.32} \\
0 & 70 & GT & C2 & 97.14 & 99.02 & 99.56 & 98.35 & 99.05 \\
\addlinespace[0.5pt]
0 & 100 & GT & C1 & 99.27 & 99.71 & 99.84 & 99.40 & 99.66 \\
0 & 100 & GT & C2 & \textbf{99.59} & \textbf{99.81} & \textbf{99.94} & \textbf{99.76} & \textbf{99.87} \\
\addlinespace[0.5pt]
0 & 30 & N & C1 & \textbf{99.08} & 99.65 & \textbf{99.81} & \textbf{99.02} & \textbf{99.45} \\
0 & 30 & N & C2 & 98.06 & \textbf{99.68} & 99.78 & 98.85 & 99.33 \\
\addlinespace[0.5pt]
0 & 50 & N & C1 & \textbf{98.92} & \textbf{99.68} & \textbf{99.81} & \textbf{98.98} & \textbf{99.43} \\
0 & 50 & N & C2 & 97.78 & 99.56 & 99.78 & 98.74 & 99.27 \\
\addlinespace[0.5pt]
0 & 70 & N & C1 & \textbf{97.59} & \textbf{99.27} & \textbf{99.84} & \textbf{98.74} & \textbf{99.31} \\
0 & 70 & N & C2 & 97.21 & 99.02 & 99.56 & 98.35 & 99.05 \\
\addlinespace[0.5pt]
0 & 100 & N & C1 & 99.24 & 99.68 & 99.84 & 99.39 & 99.66 \\
0 & 100 & N & C2 & \textbf{99.59} & \textbf{99.81} & \textbf{99.94} & \textbf{99.72} & \textbf{99.85} \\
\addlinespace[0.5pt]
100 & 100 & GT & C1 & 98.73 & 99.75 & 99.87 & 99.17 & 99.54 \\
100 & 100 & GT & C2 & \textbf{99.27} & \textbf{99.81} & \textbf{99.94} & \textbf{99.64} & \textbf{99.79} \\
\addlinespace[0.5pt]
100 & 100 & N & C1 & 98.63 & 99.75 & 99.87 & 99.13 & 99.52 \\
100 & 100 & N & C2 & \textbf{99.27} & \textbf{99.78} & \textbf{99.94} & \textbf{99.60} & \textbf{99.77} \\
\addlinespace[0.5pt]
\bottomrule
\end{tabular*}
\end{minipage}\hfill
\begin{minipage}[t]{0.49\linewidth}
\centering
\textbf{(c) Per-frame median-aligned depth}\par\smallskip
\begin{tabular*}{\linewidth}{@{\extracolsep{\fill}}rrclrrrr@{}}
\toprule
D & C & Pose & Model & AbsRel$\downarrow$ & $\delta_1\uparrow$ & AbsRel-U$\downarrow$ & $\delta_{1,U}\uparrow$ \\
\midrule
0 & 0 & -- & C1 & 0.080 & \textbf{92.15} & 0.080 & \textbf{92.15} \\
0 & 0 & -- & C2 & 0.080 & 92.09 & 0.080 & 92.09 \\
\addlinespace[0.5pt]
30 & 0 & -- & C1 & 0.071 & \textbf{93.37} & 0.077 & \textbf{92.59} \\
30 & 0 & -- & C2 & 0.071 & 93.33 & 0.077 & 92.53 \\
\addlinespace[0.5pt]
50 & 0 & -- & C1 & 0.067 & \textbf{93.95} & 0.077 & \textbf{92.66} \\
50 & 0 & -- & C2 & \textbf{0.066} & 93.93 & 0.077 & 92.61 \\
\addlinespace[0.5pt]
70 & 0 & -- & C1 & 0.062 & \textbf{94.51} & 0.077 & \textbf{92.64} \\
70 & 0 & -- & C2 & 0.062 & 94.49 & \textbf{0.076} & 92.60 \\
\addlinespace[0.5pt]
100 & 0 & -- & C1 & 0.056 & 95.39 & -- & -- \\
100 & 0 & -- & C2 & \textbf{0.055} & 95.39 & -- & -- \\
\addlinespace[0.5pt]
0 & 30 & GT & C1 & 0.080 & \textbf{92.22} & 0.080 & \textbf{92.22} \\
0 & 30 & GT & C2 & \textbf{0.079} & 92.16 & \textbf{0.079} & 92.16 \\
\addlinespace[0.5pt]
0 & 50 & GT & C1 & 0.080 & \textbf{92.22} & 0.080 & \textbf{92.22} \\
0 & 50 & GT & C2 & \textbf{0.079} & 92.17 & \textbf{0.079} & 92.17 \\
\addlinespace[0.5pt]
0 & 70 & GT & C1 & 0.080 & \textbf{92.23} & 0.080 & \textbf{92.23} \\
0 & 70 & GT & C2 & \textbf{0.079} & 92.17 & \textbf{0.079} & 92.17 \\
\addlinespace[0.5pt]
0 & 100 & GT & C1 & 0.080 & \textbf{92.23} & 0.080 & \textbf{92.23} \\
0 & 100 & GT & C2 & \textbf{0.079} & 92.17 & \textbf{0.079} & 92.17 \\
\addlinespace[0.5pt]
0 & 30 & N & C1 & 0.080 & \textbf{92.22} & 0.080 & \textbf{92.22} \\
0 & 30 & N & C2 & \textbf{0.079} & 92.16 & \textbf{0.079} & 92.16 \\
\addlinespace[0.5pt]
0 & 50 & N & C1 & 0.080 & \textbf{92.22} & 0.080 & \textbf{92.22} \\
0 & 50 & N & C2 & \textbf{0.079} & 92.17 & \textbf{0.079} & 92.17 \\
\addlinespace[0.5pt]
0 & 70 & N & C1 & 0.080 & \textbf{92.23} & 0.080 & \textbf{92.23} \\
0 & 70 & N & C2 & \textbf{0.079} & 92.17 & \textbf{0.079} & 92.17 \\
\addlinespace[0.5pt]
0 & 100 & N & C1 & 0.080 & \textbf{92.23} & 0.080 & \textbf{92.23} \\
0 & 100 & N & C2 & \textbf{0.079} & 92.17 & \textbf{0.079} & 92.17 \\
\addlinespace[0.5pt]
100 & 100 & GT & C1 & 0.056 & 95.40 & -- & -- \\
100 & 100 & GT & C2 & \textbf{0.055} & 95.40 & -- & -- \\
\addlinespace[0.5pt]
100 & 100 & N & C1 & 0.056 & 95.40 & -- & -- \\
100 & 100 & N & C2 & \textbf{0.055} & 95.40 & -- & -- \\
\addlinespace[0.5pt]
\bottomrule
\end{tabular*}
\end{minipage}
\par\smallskip
R/T/A denote RRA/RTA/AUC; panel (b) uses directed errors. U denotes unprovided depth views. D/C are provided-view percentages.
\endgroup
\end{table}

\begin{table}[p]
\centering
\caption{\textbf{Complete matched-training results on TUM RGB-D.} C1: OmniVGGT-trained; C2: VPGGT (PRC only). Bold marks the better displayed score within each matched pair. Angular thresholds are in degrees; angular scores and $\delta_1$ are percentages.}
\label{tab:trained-tum-eval-compact}
\begingroup\scriptsize
\setlength{\tabcolsep}{1.1pt}
\renewcommand{\arraystretch}{0.90}
\textbf{(a) Sign-free pose metrics}\par\smallskip
\begin{tabular*}{\linewidth}{@{\extracolsep{\fill}}rrclrrrrrrrr@{}}
\toprule
D & C & Pose & Model & R@2$\uparrow$ & R@5$\uparrow$ & R@15$\uparrow$ & T@2$\uparrow$ & T@5$\uparrow$ & T@15$\uparrow$ & A@15$\uparrow$ & A@30$\uparrow$ \\
\midrule
0 & 0 & -- & C1 & 94.78 & 96.33 & 97.22 & \textbf{22.22} & \textbf{51.22} & 75.78 & \textbf{56.78} & 69.17 \\
0 & 0 & -- & C2 & \textbf{95.44} & \textbf{97.22} & 97.22 & 20.22 & 50.33 & \textbf{76.56} & 56.53 & \textbf{69.33} \\
\addlinespace[0.5pt]
30 & 0 & -- & C1 & 94.11 & 97.00 & 97.22 & \textbf{17.44} & 47.33 & 73.00 & 52.87 & 66.46 \\
30 & 0 & -- & C2 & \textbf{96.56} & \textbf{97.22} & 97.22 & 17.11 & \textbf{48.56} & \textbf{73.44} & \textbf{52.90} & \textbf{66.66} \\
\addlinespace[0.5pt]
50 & 0 & -- & C1 & 94.78 & 95.56 & 97.22 & \textbf{17.11} & 45.78 & \textbf{73.56} & \textbf{52.44} & 66.21 \\
50 & 0 & -- & C2 & \textbf{96.89} & \textbf{97.22} & 97.22 & 16.56 & \textbf{45.89} & 73.33 & 52.10 & \textbf{66.46} \\
\addlinespace[0.5pt]
70 & 0 & -- & C1 & 94.33 & 95.33 & 97.00 & 19.11 & 47.89 & 75.56 & 54.80 & 68.22 \\
70 & 0 & -- & C2 & \textbf{95.44} & \textbf{97.22} & \textbf{97.22} & 19.11 & \textbf{49.22} & \textbf{76.78} & \textbf{55.41} & \textbf{68.87} \\
\addlinespace[0.5pt]
100 & 0 & -- & C1 & 93.56 & 95.44 & 95.44 & \textbf{27.22} & 51.89 & 78.11 & 58.50 & 70.71 \\
100 & 0 & -- & C2 & \textbf{95.89} & \textbf{96.22} & \textbf{97.22} & 26.00 & \textbf{52.33} & \textbf{79.44} & \textbf{59.19} & \textbf{71.77} \\
\addlinespace[0.5pt]
0 & 30 & GT & C1 & 95.00 & 96.44 & 97.22 & 19.11 & 49.00 & 75.33 & 55.01 & 67.92 \\
0 & 30 & GT & C2 & \textbf{95.67} & \textbf{97.22} & 97.22 & \textbf{20.33} & \textbf{51.00} & \textbf{76.33} & \textbf{56.64} & \textbf{69.29} \\
\addlinespace[0.5pt]
0 & 50 & GT & C1 & 95.11 & 96.44 & 97.22 & 16.89 & 49.67 & 74.44 & 54.41 & 67.19 \\
0 & 50 & GT & C2 & \textbf{95.67} & \textbf{97.22} & 97.22 & \textbf{24.00} & \textbf{55.00} & \textbf{77.89} & \textbf{58.95} & \textbf{70.74} \\
\addlinespace[0.5pt]
0 & 70 & GT & C1 & 95.11 & 96.11 & 96.56 & 19.22 & 51.56 & 75.56 & 56.17 & 68.26 \\
0 & 70 & GT & C2 & \textbf{95.44} & \textbf{96.56} & \textbf{97.67} & \textbf{31.56} & \textbf{60.11} & \textbf{80.22} & \textbf{62.79} & \textbf{73.50} \\
\addlinespace[0.5pt]
0 & 100 & GT & C1 & 96.22 & \textbf{97.22} & 97.22 & 25.11 & 58.78 & 77.56 & 60.35 & 71.04 \\
0 & 100 & GT & C2 & \textbf{96.44} & 96.89 & \textbf{98.89} & \textbf{52.00} & \textbf{73.78} & \textbf{84.00} & \textbf{72.22} & \textbf{80.46} \\
\addlinespace[0.5pt]
0 & 30 & N & C1 & 95.00 & 96.44 & 97.22 & \textbf{18.67} & 49.22 & 74.67 & 54.95 & 67.85 \\
0 & 30 & N & C2 & \textbf{95.56} & \textbf{97.22} & 97.22 & 18.33 & \textbf{49.78} & \textbf{75.67} & \textbf{55.70} & \textbf{68.64} \\
\addlinespace[0.5pt]
0 & 50 & N & C1 & 95.11 & 96.44 & 97.22 & 16.22 & 49.11 & 74.56 & 54.08 & 67.00 \\
0 & 50 & N & C2 & \textbf{95.67} & \textbf{97.22} & 97.22 & \textbf{20.11} & \textbf{51.11} & \textbf{77.00} & \textbf{56.81} & \textbf{69.28} \\
\addlinespace[0.5pt]
0 & 70 & N & C1 & 95.11 & 96.11 & 96.56 & 19.67 & 51.44 & 75.56 & 56.23 & 68.25 \\
0 & 70 & N & C2 & \textbf{95.44} & \textbf{96.56} & \textbf{97.67} & \textbf{22.56} & \textbf{55.00} & \textbf{78.11} & \textbf{58.86} & \textbf{70.89} \\
\addlinespace[0.5pt]
0 & 100 & N & C1 & 96.11 & 97.22 & 97.22 & 27.22 & 58.89 & 76.11 & 60.44 & 70.89 \\
0 & 100 & N & C2 & \textbf{96.78} & 97.22 & \textbf{98.89} & \textbf{35.33} & \textbf{67.00} & \textbf{81.44} & \textbf{66.07} & \textbf{76.37} \\
\addlinespace[0.5pt]
100 & 100 & GT & C1 & 94.89 & 95.44 & 95.44 & 27.89 & 55.89 & 78.00 & 59.99 & 71.86 \\
100 & 100 & GT & C2 & \textbf{95.33} & \textbf{96.56} & \textbf{97.22} & \textbf{41.78} & \textbf{68.44} & \textbf{84.78} & \textbf{68.91} & \textbf{79.08} \\
\addlinespace[0.5pt]
100 & 100 & N & C1 & 94.78 & 95.44 & 95.44 & 27.33 & 55.56 & 78.56 & 59.01 & 71.16 \\
100 & 100 & N & C2 & \textbf{95.22} & \textbf{96.67} & \textbf{97.22} & \textbf{27.78} & \textbf{63.44} & \textbf{82.11} & \textbf{63.78} & \textbf{75.31} \\
\addlinespace[0.5pt]
\bottomrule
\end{tabular*}
\par\medskip
\begin{minipage}[t]{0.49\linewidth}
\centering
\textbf{(b) Directed translation and joint pose}\par\smallskip
\begin{tabular*}{\linewidth}{@{\extracolsep{\fill}}rrclrrrrr@{}}
\toprule
D & C & Pose & Model & T@2$\uparrow$ & T@5$\uparrow$ & T@15$\uparrow$ & A@15$\uparrow$ & A@30$\uparrow$ \\
\midrule
0 & 0 & -- & C1 & \textbf{22.22} & \textbf{51.22} & 75.67 & \textbf{56.73} & 69.09 \\
0 & 0 & -- & C2 & 20.22 & 50.33 & \textbf{76.44} & 56.50 & \textbf{69.17} \\
\addlinespace[0.5pt]
30 & 0 & -- & C1 & \textbf{17.44} & 47.33 & 72.78 & 52.84 & 66.23 \\
30 & 0 & -- & C2 & 17.11 & \textbf{48.56} & \textbf{73.44} & \textbf{52.90} & \textbf{66.56} \\
\addlinespace[0.5pt]
50 & 0 & -- & C1 & \textbf{17.11} & 45.78 & \textbf{73.44} & \textbf{52.42} & 66.00 \\
50 & 0 & -- & C2 & 16.56 & \textbf{45.89} & 73.22 & 52.10 & \textbf{66.39} \\
\addlinespace[0.5pt]
70 & 0 & -- & C1 & 19.11 & 47.89 & 75.33 & 54.75 & 67.91 \\
70 & 0 & -- & C2 & 19.11 & \textbf{49.22} & \textbf{76.67} & \textbf{55.39} & \textbf{68.66} \\
\addlinespace[0.5pt]
100 & 0 & -- & C1 & \textbf{27.22} & 51.89 & 78.00 & 58.50 & 70.58 \\
100 & 0 & -- & C2 & 26.00 & \textbf{52.33} & \textbf{79.33} & \textbf{59.19} & \textbf{71.65} \\
\addlinespace[0.5pt]
0 & 30 & GT & C1 & 19.11 & 49.00 & 75.33 & 55.01 & 67.88 \\
0 & 30 & GT & C2 & \textbf{20.33} & \textbf{51.00} & \textbf{76.33} & \textbf{56.64} & \textbf{69.21} \\
\addlinespace[0.5pt]
0 & 50 & GT & C1 & 16.89 & 49.67 & 74.44 & 54.41 & 67.17 \\
0 & 50 & GT & C2 & \textbf{24.00} & \textbf{55.00} & \textbf{77.67} & \textbf{58.95} & \textbf{70.63} \\
\addlinespace[0.5pt]
0 & 70 & GT & C1 & 19.22 & 51.56 & 75.44 & 56.14 & 68.19 \\
0 & 70 & GT & C2 & \textbf{31.56} & \textbf{60.11} & \textbf{80.22} & \textbf{62.79} & \textbf{73.49} \\
\addlinespace[0.5pt]
0 & 100 & GT & C1 & 25.11 & 58.78 & 77.56 & 60.35 & 71.04 \\
0 & 100 & GT & C2 & \textbf{52.00} & \textbf{73.78} & \textbf{84.00} & \textbf{72.22} & \textbf{80.46} \\
\addlinespace[0.5pt]
0 & 30 & N & C1 & \textbf{18.67} & 49.22 & 74.67 & 54.95 & 67.81 \\
0 & 30 & N & C2 & 18.33 & \textbf{49.78} & \textbf{75.67} & \textbf{55.70} & \textbf{68.53} \\
\addlinespace[0.5pt]
0 & 50 & N & C1 & 16.22 & 49.11 & 74.56 & 54.08 & 66.98 \\
0 & 50 & N & C2 & \textbf{20.11} & \textbf{51.11} & \textbf{76.78} & \textbf{56.81} & \textbf{69.21} \\
\addlinespace[0.5pt]
0 & 70 & N & C1 & 19.67 & 51.44 & 75.22 & 56.22 & 68.19 \\
0 & 70 & N & C2 & \textbf{22.56} & \textbf{55.00} & \textbf{78.11} & \textbf{58.86} & \textbf{70.87} \\
\addlinespace[0.5pt]
0 & 100 & N & C1 & 27.22 & 58.89 & 76.11 & 60.44 & 70.89 \\
0 & 100 & N & C2 & \textbf{35.33} & \textbf{67.00} & \textbf{81.44} & \textbf{66.07} & \textbf{76.37} \\
\addlinespace[0.5pt]
100 & 100 & GT & C1 & 27.89 & 55.89 & 78.00 & 59.99 & 71.81 \\
100 & 100 & GT & C2 & \textbf{41.78} & \textbf{68.44} & \textbf{84.78} & \textbf{68.91} & \textbf{79.08} \\
\addlinespace[0.5pt]
100 & 100 & N & C1 & 27.33 & 55.56 & 78.56 & 59.01 & 71.13 \\
100 & 100 & N & C2 & \textbf{27.78} & \textbf{63.44} & \textbf{82.11} & \textbf{63.78} & \textbf{75.31} \\
\addlinespace[0.5pt]
\bottomrule
\end{tabular*}
\end{minipage}\hfill
\begin{minipage}[t]{0.49\linewidth}
\centering
\textbf{(c) Per-frame median-aligned depth}\par\smallskip
\begin{tabular*}{\linewidth}{@{\extracolsep{\fill}}rrclrrrr@{}}
\toprule
D & C & Pose & Model & AbsRel$\downarrow$ & $\delta_1\uparrow$ & AbsRel-U$\downarrow$ & $\delta_{1,U}\uparrow$ \\
\midrule
0 & 0 & -- & C1 & 0.058 & \textbf{94.64} & 0.058 & \textbf{94.64} \\
0 & 0 & -- & C2 & 0.058 & 94.41 & 0.058 & 94.41 \\
\addlinespace[0.5pt]
30 & 0 & -- & C1 & 0.031 & \textbf{98.05} & 0.040 & \textbf{97.32} \\
30 & 0 & -- & C2 & 0.031 & 98.04 & 0.040 & 97.31 \\
\addlinespace[0.5pt]
50 & 0 & -- & C1 & 0.024 & \textbf{98.58} & 0.038 & \textbf{97.40} \\
50 & 0 & -- & C2 & \textbf{0.023} & 98.57 & \textbf{0.037} & 97.39 \\
\addlinespace[0.5pt]
70 & 0 & -- & C1 & 0.017 & 99.11 & 0.034 & 97.62 \\
70 & 0 & -- & C2 & \textbf{0.016} & 99.11 & \textbf{0.033} & 97.62 \\
\addlinespace[0.5pt]
100 & 0 & -- & C1 & 0.009 & 99.76 & -- & -- \\
100 & 0 & -- & C2 & 0.009 & 99.76 & -- & -- \\
\addlinespace[0.5pt]
0 & 30 & GT & C1 & \textbf{0.057} & \textbf{94.66} & \textbf{0.057} & \textbf{94.66} \\
0 & 30 & GT & C2 & 0.058 & 94.44 & 0.058 & 94.44 \\
\addlinespace[0.5pt]
0 & 50 & GT & C1 & \textbf{0.057} & \textbf{94.68} & \textbf{0.057} & \textbf{94.68} \\
0 & 50 & GT & C2 & 0.058 & 94.46 & 0.058 & 94.46 \\
\addlinespace[0.5pt]
0 & 70 & GT & C1 & \textbf{0.057} & \textbf{94.69} & \textbf{0.057} & \textbf{94.69} \\
0 & 70 & GT & C2 & 0.058 & 94.46 & 0.058 & 94.46 \\
\addlinespace[0.5pt]
0 & 100 & GT & C1 & \textbf{0.057} & \textbf{94.69} & \textbf{0.057} & \textbf{94.69} \\
0 & 100 & GT & C2 & 0.058 & 94.49 & 0.058 & 94.49 \\
\addlinespace[0.5pt]
0 & 30 & N & C1 & \textbf{0.057} & \textbf{94.66} & \textbf{0.057} & \textbf{94.66} \\
0 & 30 & N & C2 & 0.058 & 94.44 & 0.058 & 94.44 \\
\addlinespace[0.5pt]
0 & 50 & N & C1 & \textbf{0.057} & \textbf{94.68} & \textbf{0.057} & \textbf{94.68} \\
0 & 50 & N & C2 & 0.058 & 94.46 & 0.058 & 94.46 \\
\addlinespace[0.5pt]
0 & 70 & N & C1 & \textbf{0.057} & \textbf{94.69} & \textbf{0.057} & \textbf{94.69} \\
0 & 70 & N & C2 & 0.058 & 94.46 & 0.058 & 94.46 \\
\addlinespace[0.5pt]
0 & 100 & N & C1 & \textbf{0.057} & \textbf{94.70} & \textbf{0.057} & \textbf{94.70} \\
0 & 100 & N & C2 & 0.058 & 94.51 & 0.058 & 94.51 \\
\addlinespace[0.5pt]
100 & 100 & GT & C1 & 0.009 & 99.76 & -- & -- \\
100 & 100 & GT & C2 & 0.009 & 99.76 & -- & -- \\
\addlinespace[0.5pt]
100 & 100 & N & C1 & 0.009 & 99.76 & -- & -- \\
100 & 100 & N & C2 & 0.009 & 99.76 & -- & -- \\
\addlinespace[0.5pt]
\bottomrule
\end{tabular*}
\end{minipage}
\par\smallskip
R/T/A denote RRA/RTA/AUC; panel (b) uses directed errors. U denotes unprovided depth views. D/C are provided-view percentages.
\endgroup
\end{table}

\begin{table}[p]
\centering
\caption{\textbf{Complete matched-training results on C3VD.} C1: OmniVGGT-trained; C2: VPGGT (PRC only). Bold marks the better displayed score within each matched pair. Angular thresholds are in degrees; angular scores and $\delta_1$ are percentages.}
\label{tab:trained-c3vd-eval-compact}
\begingroup\scriptsize
\setlength{\tabcolsep}{1.1pt}
\renewcommand{\arraystretch}{0.90}
\textbf{(a) Sign-free pose metrics}\par\smallskip
\begin{tabular*}{\linewidth}{@{\extracolsep{\fill}}rrclrrrrrrrr@{}}
\toprule
D & C & Pose & Model & R@2$\uparrow$ & R@5$\uparrow$ & R@15$\uparrow$ & T@2$\uparrow$ & T@5$\uparrow$ & T@15$\uparrow$ & A@15$\uparrow$ & A@30$\uparrow$ \\
\midrule
0 & 0 & -- & C1 & 100.00 & 100.00 & 100.00 & \textbf{8.64} & \textbf{29.19} & 71.85 & \textbf{43.97} & \textbf{61.55} \\
0 & 0 & -- & C2 & 100.00 & 100.00 & 100.00 & 7.65 & 27.75 & \textbf{72.30} & 42.73 & 61.49 \\
\addlinespace[0.5pt]
30 & 0 & -- & C1 & 99.95 & 100.00 & 100.00 & \textbf{6.32} & 25.33 & 60.30 & 36.48 & 54.68 \\
30 & 0 & -- & C2 & \textbf{100.00} & 100.00 & 100.00 & 6.17 & \textbf{26.12} & \textbf{61.38} & \textbf{36.58} & \textbf{54.92} \\
\addlinespace[0.5pt]
50 & 0 & -- & C1 & 100.00 & 100.00 & 100.00 & 7.75 & 30.52 & 63.41 & 39.78 & 57.00 \\
50 & 0 & -- & C2 & 100.00 & 100.00 & 100.00 & \textbf{8.40} & \textbf{30.72} & \textbf{64.69} & \textbf{40.18} & \textbf{57.29} \\
\addlinespace[0.5pt]
70 & 0 & -- & C1 & 100.00 & 100.00 & 100.00 & 11.31 & \textbf{37.48} & 69.88 & 46.17 & 62.24 \\
70 & 0 & -- & C2 & 100.00 & 100.00 & 100.00 & 11.31 & 36.99 & \textbf{70.12} & \textbf{46.45} & \textbf{62.33} \\
\addlinespace[0.5pt]
100 & 0 & -- & C1 & 100.00 & 100.00 & 100.00 & 16.74 & 53.23 & 83.51 & 59.73 & 74.26 \\
100 & 0 & -- & C2 & 100.00 & 100.00 & 100.00 & \textbf{17.04} & \textbf{54.22} & \textbf{84.30} & \textbf{60.13} & \textbf{74.59} \\
\addlinespace[0.5pt]
0 & 30 & GT & C1 & 100.00 & 100.00 & 100.00 & 8.00 & 27.95 & 70.81 & 42.07 & 60.43 \\
0 & 30 & GT & C2 & 100.00 & 100.00 & 100.00 & \textbf{9.68} & \textbf{28.74} & \textbf{71.01} & \textbf{42.23} & \textbf{60.89} \\
\addlinespace[0.5pt]
0 & 50 & GT & C1 & 100.00 & 100.00 & 100.00 & 8.69 & 28.89 & 71.80 & 43.31 & 61.84 \\
0 & 50 & GT & C2 & 100.00 & 100.00 & 100.00 & \textbf{15.01} & \textbf{35.26} & \textbf{73.19} & \textbf{46.99} & \textbf{64.52} \\
\addlinespace[0.5pt]
0 & 70 & GT & C1 & 100.00 & 100.00 & 100.00 & 9.23 & 32.30 & 73.98 & 46.34 & 64.50 \\
0 & 70 & GT & C2 & 100.00 & 100.00 & 100.00 & \textbf{25.28} & \textbf{46.81} & \textbf{78.67} & \textbf{56.20} & \textbf{71.01} \\
\addlinespace[0.5pt]
0 & 100 & GT & C1 & 100.00 & 100.00 & 100.00 & 10.37 & 43.16 & 78.37 & 52.76 & 69.42 \\
0 & 100 & GT & C2 & 100.00 & 100.00 & 100.00 & \textbf{47.90} & \textbf{76.74} & \textbf{95.41} & \textbf{78.63} & \textbf{88.48} \\
\addlinespace[0.5pt]
0 & 30 & N & C1 & 100.00 & 100.00 & 100.00 & \textbf{7.90} & \textbf{27.65} & \textbf{69.78} & \textbf{41.53} & \textbf{59.62} \\
0 & 30 & N & C2 & 100.00 & 100.00 & 100.00 & 7.26 & 26.02 & 68.44 & 39.77 & 58.60 \\
\addlinespace[0.5pt]
0 & 50 & N & C1 & 100.00 & 100.00 & 100.00 & 8.30 & 28.00 & \textbf{70.62} & \textbf{42.30} & \textbf{60.38} \\
0 & 50 & N & C2 & 100.00 & 100.00 & 100.00 & \textbf{10.22} & \textbf{28.84} & 67.75 & 40.87 & 58.77 \\
\addlinespace[0.5pt]
0 & 70 & N & C1 & 100.00 & 100.00 & 100.00 & 9.19 & 31.85 & \textbf{72.49} & \textbf{45.31} & \textbf{62.94} \\
0 & 70 & N & C2 & 100.00 & 100.00 & 100.00 & \textbf{14.81} & \textbf{34.77} & 68.69 & 45.06 & 61.50 \\
\addlinespace[0.5pt]
0 & 100 & N & C1 & 100.00 & 100.00 & 100.00 & 9.48 & 42.62 & 77.38 & 51.77 & 67.73 \\
0 & 100 & N & C2 & 100.00 & 100.00 & 100.00 & \textbf{29.28} & \textbf{55.46} & \textbf{79.46} & \textbf{60.70} & \textbf{72.81} \\
\addlinespace[0.5pt]
100 & 100 & GT & C1 & 100.00 & 100.00 & 100.00 & 20.74 & 55.51 & 85.19 & 62.26 & 76.36 \\
100 & 100 & GT & C2 & 100.00 & 100.00 & 100.00 & \textbf{56.49} & \textbf{81.78} & \textbf{97.19} & \textbf{82.95} & \textbf{91.04} \\
\addlinespace[0.5pt]
100 & 100 & N & C1 & 100.00 & 100.00 & 100.00 & 19.85 & 54.07 & \textbf{83.16} & 60.86 & \textbf{74.80} \\
100 & 100 & N & C2 & 100.00 & 100.00 & 100.00 & \textbf{35.16} & \textbf{60.40} & 80.15 & \textbf{63.58} & 74.53 \\
\addlinespace[0.5pt]
\bottomrule
\end{tabular*}
\par\medskip
\begin{minipage}[t]{0.49\linewidth}
\centering
\textbf{(b) Directed translation and joint pose}\par\smallskip
\begin{tabular*}{\linewidth}{@{\extracolsep{\fill}}rrclrrrrr@{}}
\toprule
D & C & Pose & Model & T@2$\uparrow$ & T@5$\uparrow$ & T@15$\uparrow$ & A@15$\uparrow$ & A@30$\uparrow$ \\
\midrule
0 & 0 & -- & C1 & \textbf{8.64} & \textbf{29.19} & 71.70 & \textbf{43.95} & \textbf{61.39} \\
0 & 0 & -- & C2 & 7.65 & 27.75 & \textbf{72.20} & 42.70 & 61.27 \\
\addlinespace[0.5pt]
30 & 0 & -- & C1 & \textbf{6.32} & 25.28 & 59.80 & 36.30 & 54.17 \\
30 & 0 & -- & C2 & 6.17 & \textbf{26.07} & \textbf{60.99} & \textbf{36.45} & \textbf{54.39} \\
\addlinespace[0.5pt]
50 & 0 & -- & C1 & 7.70 & 30.47 & 63.01 & 39.63 & 56.60 \\
50 & 0 & -- & C2 & \textbf{8.40} & \textbf{30.67} & \textbf{64.30} & \textbf{40.06} & \textbf{56.78} \\
\addlinespace[0.5pt]
70 & 0 & -- & C1 & 11.31 & \textbf{37.43} & 69.58 & 46.08 & 61.89 \\
70 & 0 & -- & C2 & 11.31 & 36.99 & \textbf{69.83} & \textbf{46.35} & 61.89 \\
\addlinespace[0.5pt]
100 & 0 & -- & C1 & 16.74 & 53.23 & 83.51 & 59.73 & 74.26 \\
100 & 0 & -- & C2 & \textbf{17.04} & \textbf{54.22} & \textbf{84.30} & \textbf{60.13} & \textbf{74.58} \\
\addlinespace[0.5pt]
0 & 30 & GT & C1 & 8.00 & 27.95 & 70.72 & 42.05 & 60.29 \\
0 & 30 & GT & C2 & \textbf{9.68} & \textbf{28.74} & \textbf{70.91} & \textbf{42.19} & \textbf{60.70} \\
\addlinespace[0.5pt]
0 & 50 & GT & C1 & 8.69 & 28.89 & 71.70 & 43.29 & 61.68 \\
0 & 50 & GT & C2 & \textbf{15.01} & \textbf{35.26} & \textbf{73.14} & \textbf{46.99} & \textbf{64.41} \\
\addlinespace[0.5pt]
0 & 70 & GT & C1 & 9.23 & 32.25 & 73.88 & 46.28 & 64.36 \\
0 & 70 & GT & C2 & \textbf{25.28} & \textbf{46.81} & \textbf{78.62} & \textbf{56.17} & \textbf{70.91} \\
\addlinespace[0.5pt]
0 & 100 & GT & C1 & 10.37 & 43.16 & 78.37 & 52.76 & 69.39 \\
0 & 100 & GT & C2 & \textbf{47.90} & \textbf{76.74} & \textbf{95.41} & \textbf{78.63} & \textbf{88.48} \\
\addlinespace[0.5pt]
0 & 30 & N & C1 & \textbf{7.85} & \textbf{27.60} & \textbf{69.63} & \textbf{41.47} & \textbf{59.40} \\
0 & 30 & N & C2 & 7.26 & 26.02 & 68.25 & 39.70 & 58.35 \\
\addlinespace[0.5pt]
0 & 50 & N & C1 & 8.30 & 27.95 & \textbf{70.52} & \textbf{42.26} & \textbf{60.22} \\
0 & 50 & N & C2 & \textbf{10.22} & \textbf{28.84} & 67.65 & 40.85 & 58.59 \\
\addlinespace[0.5pt]
0 & 70 & N & C1 & 9.14 & 31.80 & \textbf{72.40} & \textbf{45.23} & \textbf{62.75} \\
0 & 70 & N & C2 & \textbf{14.81} & \textbf{34.77} & 68.64 & 45.05 & 61.35 \\
\addlinespace[0.5pt]
0 & 100 & N & C1 & 9.48 & 42.57 & 77.33 & 51.72 & 67.66 \\
0 & 100 & N & C2 & \textbf{29.28} & \textbf{55.46} & \textbf{79.36} & \textbf{60.67} & \textbf{72.75} \\
\addlinespace[0.5pt]
100 & 100 & GT & C1 & 20.74 & 55.51 & 85.14 & 62.26 & 76.34 \\
100 & 100 & GT & C2 & \textbf{56.49} & \textbf{81.78} & \textbf{97.19} & \textbf{82.95} & \textbf{91.04} \\
\addlinespace[0.5pt]
100 & 100 & N & C1 & 19.85 & 54.07 & \textbf{83.16} & 60.86 & \textbf{74.78} \\
100 & 100 & N & C2 & \textbf{35.16} & \textbf{60.40} & 80.05 & \textbf{63.56} & 74.47 \\
\addlinespace[0.5pt]
\bottomrule
\end{tabular*}
\end{minipage}\hfill
\begin{minipage}[t]{0.49\linewidth}
\centering
\textbf{(c) Per-frame median-aligned depth}\par\smallskip
\begin{tabular*}{\linewidth}{@{\extracolsep{\fill}}rrclrrrr@{}}
\toprule
D & C & Pose & Model & AbsRel$\downarrow$ & $\delta_1\uparrow$ & AbsRel-U$\downarrow$ & $\delta_{1,U}\uparrow$ \\
\midrule
0 & 0 & -- & C1 & 0.055 & 97.87 & 0.055 & 97.87 \\
0 & 0 & -- & C2 & 0.055 & \textbf{98.09} & 0.055 & \textbf{98.09} \\
\addlinespace[0.5pt]
30 & 0 & -- & C1 & 0.015 & 99.78 & 0.020 & 99.70 \\
30 & 0 & -- & C2 & 0.015 & \textbf{99.80} & \textbf{0.019} & \textbf{99.72} \\
\addlinespace[0.5pt]
50 & 0 & -- & C1 & 0.011 & 99.86 & 0.017 & 99.75 \\
50 & 0 & -- & C2 & \textbf{0.010} & \textbf{99.87} & 0.017 & 99.75 \\
\addlinespace[0.5pt]
70 & 0 & -- & C1 & 0.007 & 99.92 & 0.015 & 99.77 \\
70 & 0 & -- & C2 & 0.007 & 99.92 & \textbf{0.014} & \textbf{99.78} \\
\addlinespace[0.5pt]
100 & 0 & -- & C1 & 0.004 & 99.98 & -- & -- \\
100 & 0 & -- & C2 & 0.004 & 99.98 & -- & -- \\
\addlinespace[0.5pt]
0 & 30 & GT & C1 & 0.055 & 97.88 & 0.055 & 97.88 \\
0 & 30 & GT & C2 & 0.055 & \textbf{98.10} & 0.055 & \textbf{98.10} \\
\addlinespace[0.5pt]
0 & 50 & GT & C1 & 0.055 & 97.89 & 0.055 & 97.89 \\
0 & 50 & GT & C2 & 0.055 & \textbf{98.11} & 0.055 & \textbf{98.11} \\
\addlinespace[0.5pt]
0 & 70 & GT & C1 & 0.055 & 97.89 & 0.055 & 97.89 \\
0 & 70 & GT & C2 & 0.055 & \textbf{98.11} & 0.055 & \textbf{98.11} \\
\addlinespace[0.5pt]
0 & 100 & GT & C1 & 0.055 & 97.90 & 0.055 & 97.90 \\
0 & 100 & GT & C2 & 0.055 & \textbf{98.11} & 0.055 & \textbf{98.11} \\
\addlinespace[0.5pt]
0 & 30 & N & C1 & 0.055 & 97.88 & 0.055 & 97.88 \\
0 & 30 & N & C2 & 0.055 & \textbf{98.10} & 0.055 & \textbf{98.10} \\
\addlinespace[0.5pt]
0 & 50 & N & C1 & 0.055 & 97.89 & 0.055 & 97.89 \\
0 & 50 & N & C2 & 0.055 & \textbf{98.11} & 0.055 & \textbf{98.11} \\
\addlinespace[0.5pt]
0 & 70 & N & C1 & 0.055 & 97.89 & 0.055 & 97.89 \\
0 & 70 & N & C2 & 0.055 & \textbf{98.11} & 0.055 & \textbf{98.11} \\
\addlinespace[0.5pt]
0 & 100 & N & C1 & 0.055 & 97.90 & 0.055 & 97.90 \\
0 & 100 & N & C2 & 0.055 & \textbf{98.11} & 0.055 & \textbf{98.11} \\
\addlinespace[0.5pt]
100 & 100 & GT & C1 & 0.004 & 99.98 & -- & -- \\
100 & 100 & GT & C2 & 0.004 & 99.98 & -- & -- \\
\addlinespace[0.5pt]
100 & 100 & N & C1 & 0.004 & 99.98 & -- & -- \\
100 & 100 & N & C2 & 0.004 & 99.98 & -- & -- \\
\addlinespace[0.5pt]
\bottomrule
\end{tabular*}
\end{minipage}
\par\smallskip
R/T/A denote RRA/RTA/AUC; panel (b) uses directed errors. U denotes unprovided depth views. D/C are provided-view percentages.
\endgroup
\end{table}

\subsection{PRC Implementation Details}
\label{app:prc_details}

PRC operates on the camera head's pre-activation nine-dimensional
state.
For availability indicators $m_i$, the translation norm ratio is
\begin{equation}
\alpha =
\frac{\sum_i m_i\|\mathbf r_{i,t}^{0}\|_2}
     {\sum_i m_i\|\bar{\mathbf t}_i^p\|_2}.
\end{equation}
The implementation applies this ratio only when the prior
translation-norm sum exceeds $10^{-4}$, with a denominator floor
of $10^{-6}$.
Otherwise, it uses $\alpha=1$.
The aligned prior is
$\mathbf a_i=(\alpha\bar{\mathbf t}_i^p,
\mathbf q_i^p,\mathbf f_i^p)$.

The IT0 prediction is recorded before state replacement.
Provided views then use $\mathbf a_i$, while unprovided views
retain their IT0 states.
The running state is detached before further refinement, following
the inherited camera-head computation.
Subsequent updates are additive in the camera encoding rather
than compositions on SE(3).
The existing activation and camera-decoding functions are retained.

PRC does not add a new refinement network or introduce the original
residual-update mechanism.
It changes the state to which that mechanism is applied.
For a provided view, the numerical forward computation can be
written as
$\mathbf u_i^{J-1}
=\mathbf a_i+\sum_{k=1}^{J-1}\Delta_i^k$.
This expression describes forward values, not an uninterrupted
gradient path through the iterations.
The residuals are unconstrained, so zero corrections preserve
the aligned prior state but are not enforced.

\subsection{Further Directions}
\label{app:future_directions}

\paragraph{Raw depth priors.}
This work uses dataset depth references without additional corruption.
Extending the model to raw depth observations requires studying
sensor-specific noise, missing values, and systematic artifacts.
A dedicated depth-processing front end is one possible extension,
but is not evaluated here.

\paragraph{Additional sensing modalities.}
Other sensors may contribute through upstream estimators that produce
camera or depth observations.
Such integration requires appropriate calibration, coordinate
conversion, and treatment of uncertainty.
The current geometric interface does not by itself establish
robustness to arbitrary sensor error distributions.

\paragraph{Prior reliability.}
Detecting and rejecting unreliable priors is an important direction.
The present model does not include an explicit reliability gate,
and its results do not establish safe fallback behavior under
severe prior corruption.

\paragraph{Vision dropout.}
The current architecture requires images.
Training and evaluating missing or degraded visual inputs would
require additional design and supervision.
Sensor-only inference and fallback to odometry are not demonstrated
in this work.

\paragraph{Streaming reconstruction.}
Extending window-level prior conditioning to streaming reconstruction
may enable longer-term geometric constraints.
Maintaining a consistent state across windows and incorporating
loop-closure information remain open integration problems.

\end{document}